\documentclass{article}

\usepackage{microtype}
\usepackage{graphicx}
\usepackage{subfigure}
\usepackage{booktabs} 

\usepackage{hyperref}

\usepackage[accepted]{icml2024}

\usepackage{amsmath}
\usepackage{amssymb}
\usepackage{mathtools}
\usepackage{amsthm}

\usepackage{caption}
\usepackage{wrapfig}
\usepackage{multicol}
\usepackage{subcaption}

\usepackage[capitalize,noabbrev]{cleveref}

\theoremstyle{plain}
\newtheorem{theorem}{Theorem}[section]

\newtheorem{lemma}[theorem]{Lemma}

\theoremstyle{definition}

\theoremstyle{remark}

\usepackage[textsize=tiny]{todonotes}

\icmltitlerunning{Variance reduction of diffusion model's gradients with Taylor approximation-based control variate}

\begin{document}

\twocolumn[
\icmltitle{Variance reduction of diffusion model's gradients with Taylor approximation-based control variate}




\begin{icmlauthorlist}
\icmlauthor{Paul Jeha}{dtu}
\icmlauthor{Will Grathwohl}{deepmind}
\icmlauthor{Michael Riis Andersen}{dtu}
\icmlauthor{Carl Henrik Ek}{cambridge}
\icmlauthor{Jes Frellsen}{dtu}
\end{icmlauthorlist}

\icmlaffiliation{dtu}{Technical University of Denmark}
\icmlaffiliation{deepmind}{Google Deepmind}
\icmlaffiliation{cambridge}{University of Cambridge}

\icmlcorrespondingauthor{Paul Jeha}{pauje@dtu.dk}

\icmlkeywords{Machine Learning, ICML}

\vskip 0.3in
]



\printAffiliationsAndNotice{} 

\begin{abstract}
Score-based models, trained with denoising score matching, are remarkably effective in generating high dimensional data. However, the high variance of their training objective hinders optimisation. We attempt to reduce it with a control variate, derived via a \(k\)-th order Taylor expansion on the training objective and its gradient. We prove an equivalence between the two and demonstrate empirically the effectiveness of our approach on a low dimensional problem setting; and study its effect on larger problems.
\end{abstract}

\section{Introduction}
\label{sec:intro}

In the field of probabilistic generative models, we find several established methods to model unknown data distribution, such as Variational Auto-Encoders \citep[VAEs;][]{Kingma_2019, 6795935}, Energy-Based Models \citep[EBMs;][]{928a56b7d6f1473e930f282a0c4b534e, grathwohlthesis, xie2022tale, du2023reduce} and Normalising Flows \citep{papamakarios2021normalizing}. Each of these methods has been designed to model and maximise the log-likelihood of the data. However, direct optimisation of the log-density incurs important constraints on the design of these models: VAEs maximise a lower bound (ELBO) of the log-likelihood, a bound that is often not tight \citep{rainforth2019tighter}. EBMs address the challenging task of estimating the partition function of the density, and Normalising Flow can only train specialised neural network architecture for which the inverse can be computed.  Score-based models emerge as an attractive alternative that circumvents those challenges by modelling the Stein score of the log-density, that is, the gradient of the log-density \citep{JMLR:v6:hyvarinen05a, 6795935}. In addition, it has been shown that training score-based models is equivalent, under certain assumptions, to maximise the log-likelihood of the data \citep{huang2021variational, song2021maximum}.

In practice, score-based models require no specialised architecture and are trained via a score matching loss, such as sliced score matching and denoising score matching \citep{song2019sliced, song2021scorebased}. Denoising score matching is a technique similar to denoising diffusion probabilistic models \citep{ho2020denoising, luo2022understanding}, where the data is corrupted with a varying amount of noise and a denoiser is trained to recover the signal from the corrupted data. While very effective, this solution suffers from high variance, making optimisation challenging \citep{song2021train}.  We propose to use a popular variance reduction method, control variate \citep{mcbook}, to address this high variance. Control variate reduces the variance by leveraging an auxiliary Monte Carlo integration problem that correlates with the original one. Control variate for score-based model has been originally introduced by \citet{wang2020wasserstein} through a linearisation of the training objective for small noise level.

We propose to generalise their method to $k$-th order Taylor approximation, which is designed for any noise value $\sigma$. 
Our contributions include: (1) deriving a control variate with an arbitrary order Taylor polynomial; (2) proving an equivalence between controlling the training objectives and its gradient; (3) empirically demonstrating the necessity of having a regression coefficient; (4) demonstrating the effectiveness of control variate in a low dimensional problem setting; (5) studying the impact of control variate in a high dimensional case; (6) showing the limitation of Taylor based control variate.

\section{Related work}

\paragraph{Score matching} \citet{JMLR:v6:hyvarinen05a} originally introduced score matching as a method to train EBMs \citep{928a56b7d6f1473e930f282a0c4b534e, grathwohlthesis} through their Stein score, that is, the gradient of their log-density. Modelling the Stein score elegantly circumvented the need to approximate the normalization constant, a notorious challenge in the EBM literature \citep{grathwohlthesis}. The central idea in score matching is that aligning the model's gradients with those of the data is sufficient to learn a model from which we can sample from.  Different variants of that idea exist with the most notable ones being implicit score matching \citep{JMLR:v6:hyvarinen05a, NIPS2010_6f3e29a3, martens2012estimating}, sliced score matching \citep{song2019sliced} and denoising score matching \citep{6795935, NEURIPS2019_3001ef25, song2021scorebased}. Denoising score matching was originally introduced by \citet{6795935}. While there were initial attempts to scale it \citep{NIPS2010_6f3e29a3, martens2012estimating}, it was not until the work of \citet{NEURIPS2019_3001ef25, song2020improved, song2021scorebased} that it has successfully scaled. Their successful insight was to combine multiple denoising score-matching objectives, each with a different amount of corruption. Concurrently, diffusion models emerged \citep{ho2020denoising, yang2023diffusion} as an equivalent method to score-based models. Together, they have successfully been applied to various data modalities of very high dimensions \citep{Rombach_2022_CVPR, xu2022geodiff, austin2023structured, harvey2022flexible}. In addition to this empirical success,  \citet{song2021maximum, huang2021variational, albergo2023stochastic} have laid a theoretical foundation for this learning procedure exhibiting profound links to the variational framework and to stochastic and ordinary differential equations.

\paragraph{Control variate} Control variate \citep{mcbook} is a variance reduction technique for Monte Carlo integration problems that has been popular in various fields, such as in variational inference \citep{Blei_2017}. \citet{pmlr-v33-ranganath14} use control variate to reduce the variance of variational objectives; in VI, \citet{NIPS2017_325995af} mitigate the variance of reparameterization gradients estimator with control variate, hence providing more reliable gradient and getting faster and more stable convergence. In a similar vein, \citet{grathwohl2018backpropagation} propose to control the variance of gradients through a surrogate neural network, in which its own gradients act as a control variate. Building on this idea, \citet{boustati2020amortized} learn a linear control variate to control deep Gaussian processes' variance. \citet{geffner2020using} offer a comprehensive review of control variate for VI. In addition to VI, control variate is a popular tool in reinforcement learning, such as controlling the gradient in the REINFORCE algorithm \citep{Williams:92}, in advantage actor-critic \citep{mnih2016asynchronous} or in policy optimisation \cite{liu2018actiondepedent}.

\section{Theory}

Suppose an unknown data distribution $p_{\text{data}}(\mathbf{x})$ and a dataset consisting of i.i.d.\@ samples $\{ \mathbf{x}_i \in \mathbb{R}^D\}_{i=1}^N$, sampled from $p_{\text{data}}$. The Stein score, $s: \mathbb{R}^D \rightarrow \mathbb{R}^D$, $s(\mathbf{x}) = \nabla_{\mathbf{x}} \log p_{\text{data}}(\mathbf{x})$, maps a data point to the gradient field of its log-density; and it is sufficient to model it to asymptotically and approximately sample from $p_{\text{data}}$ using, e.g., Langevin based methods \citep{NEURIPS2019_3001ef25,JMLR:v6:hyvarinen05a}. We use a neural network  $s_{\mathbf{\theta}}$ to model the Stein score, parameterised by $\boldsymbol{\theta}$, and train it with denoising score matching, where we learn the score of a corrupted version of the original data distribution \citep{6795935}. 
\subsection{Denoising score matching}

We follow the approach of \citet{NEURIPS2019_3001ef25} to learn our score network $s_{\boldsymbol{\theta}}$, parameterised by $\boldsymbol{\theta} \in \mathbb{R}^p$ using a weighted denoising score matching objective $\mathcal{L}_{\boldsymbol{\theta}}(\mathbf{z}, \mathbf{x}, \Sigma)$, and refer the reader to their work for the derivation of the training objective:
\begin{equation}
\label{eq:loss}
\begin{aligned}
    &\mathcal{L}_{\boldsymbol{\theta}}(\mathbf{z}, \mathbf{x}, \Sigma) \\
    &= \mathbb{E}_{\sigma \sim \mathcal{U}\left(\Sigma\right)}\mathbb{E}_{p_{\text{data}}(\mathbf{x})}\mathbb{E}_{\mathbf{z} \sim \mathcal{N}(0, \mathbf{I}_D)} \left[ \lambda(\sigma)L_{\boldsymbol{\theta}}(\mathbf{z}, \mathbf{x}, \sigma)\right] ,
\end{aligned}
\end{equation}
where
\begin{align}
\label{eq:loss}
L_{\boldsymbol{\theta}}(\mathbf{z}, \mathbf{x}, \sigma) = \frac{1}{2} \left\| \frac{\mathbf{z}}{\sigma} + s_{\boldsymbol{\theta}}(\mathbf{x} + \sigma \mathbf{z})\right\|^2
\end{align}
and $\Sigma = \left\{\sigma_i\right\}_{j=1}^L$ is an increasing geometric sequence, $\sigma \sim \mathcal{U}\left(\Sigma\right)$ is uniformly sampled from the sequence, and $\lambda$ is a positive function such that $\lambda(\sigma)L_{\boldsymbol{\theta}}(\mathbf{z}, \mathbf{x}, \sigma)$ has approximately a constant magnitude across the different noise levels. 
This training objective, unfortunately, suffers from high variance \citep{song2021maximum, song2021train, wang2020wasserstein}, which hinders the optimisation process. We aim to reduce the variance of the Monte Carlo estimator by constructing a control variate of that estimator.

\subsection{Control variate}
\label{sec:cv_method}

Control variate \citep{mcbook} is a technique to reduce the variance of an estimator $\boldsymbol{\hat{\mu}} = (1/N)\sum_{i=1}^N L(\mathbf{z}_i)$ of a Monte Carlo integration problem, $\boldsymbol{\mu} = \mathbb{E}_{\mathbf{z}}\left[L(\mathbf{z})\right]$, by using a similar known problem, $\boldsymbol{\gamma} = \mathbb{E}_{\mathbf{z}}\left[C(\mathbf{z})\right]$, where $C$ is the control variate.
Using the control variate, we construct an equivalent integration problem in \cref{eq:mc_cv} and its \textit{regression estimator} $\boldsymbol{\hat{\mu}}_{\text{CV}, \beta}$,
\noindent
\begin{equation}
\label{eq:mc_cv}
\boldsymbol{\mu} = \mathbb{E}_{\mathbf{z}}\left[L(\mathbf{z}) - \beta C(\mathbf{z})\right] + \beta \boldsymbol{\gamma}
\end{equation}
\begin{equation}
\label{eq:reg_es}
\boldsymbol{\hat{\mu}}_{\text{CV}, \beta} = \frac{1}{N}\sum_{i=1}^N \left[L(\mathbf{z}_i) - \beta C(\mathbf{z}_i)\right] + \beta \boldsymbol{\gamma}
\end{equation}

where $\beta$ is the regression coefficient and controls the scale of the control variate. When $\beta = 0$,  $\boldsymbol{\hat{\mu}}_{\text{CV}, \beta}$ equals the original estimator $\boldsymbol{\hat{\mu}}$. For any $\beta$, $\boldsymbol{\hat{\mu}}_{\text{CV}, \beta}$ is an unbiased estimator, that is $\mathbb{E}_{\mathbf{z}_1, \ldots, \mathbf{z}_N}\left[\boldsymbol{\hat{\mu}}_{\text{CV}, \beta}\right] = \mathbb{E}_{\mathbf{z}_1, \ldots, \mathbf{z}_N}\left[\boldsymbol{\hat{\mu}}\right] = \boldsymbol{\mu}$ for all $N$. There exists an optimal value $\beta_{\text{opt}}$ for which the reduction in variance is maximised. To obtain it, we derive first the variance of $ \boldsymbol{\hat{\mu}}_{\text{CV}, \beta}$:
\begin{equation}
\label{eq:var_cv}
\begin{aligned}
    \mathrm{Var}(\boldsymbol{\hat{\mu}}_{\text{CV}, \beta}) &= \frac{1}{N} \left( \mathrm{Var}(L(\mathbf{z})) - 2\beta\mathrm{Cov}(L(\mathbf{z}), C(\mathbf{z})) \right. \\
    &\left.+ \beta^2\mathrm{Var}(C(\mathbf{z})) \right)
\end{aligned}
\end{equation}
By differentiating this expression with respect to $\beta$, and zeroing it, we find the optimal value, $\beta_{\text{opt}} = \operatorname{Cov}(L(\mathbf{z}), C(\mathbf{z}))/\operatorname{Var(C(\mathbf{z}))}$, for which the variance of $\boldsymbol{\hat{\mu}}_{\text{CV}, \beta_{\text{opt}}}$ is minimal. Intuitively, the ``best'' control variate $C$ equals the original function $L$, resulting in $\beta_{\text{opt}} = 1$. To gain more intuition into what a good control variate is, we substitute  $\beta_{\text{opt}}$ in \cref{eq:var_cv}, yielding the variance of the regression estimator
\begin{equation}
\label{eq:var_cv_opt}
    \operatorname{Var}\left[ \boldsymbol{\hat{\mu}}_{\text{CV}, \beta_{\text{opt}}}\right] = \frac{1}{N} \operatorname{Var}\left[ \boldsymbol{\hat{\mu}}\right] \left( 1 - \operatorname{Corr}(L(\mathbf{z}), C(\mathbf{z}))^2\right) .
\end{equation}
\Cref{eq:var_cv_opt} shows that, given the optimal value $\beta_{\text{opt}}$, \textit{any function $C$ that correlates to $L$} (positively or negatively) reduces the variance. The main challenge is finding an appropriate function $C$. One approach, taken by \citet{wang2020wasserstein}, is to linearise the function $L$ around a point and use that as a control variate. We extend that approach by finding a suitable polynomial approximation of $L$. Various polynomial approximations exist \citep{a2ba976e-b8db-323c-8f81-cb394c5cc74f}, but one that makes sense when using automatic differentiation mechanism is the Taylor series \citep{Duistermaat2010}.

\subsection{Taylor series}

A Taylor series represents a function $s$ as a power series, whose coefficients are successive derivatives of $s$,  with an additional remainder \citep{levi1967polynomials}. Practically, a Taylor series approximates any function (cf.\@ \cref{thm:taylo}) with a polynomial, allowing control over the approximation quality through the degree of the polynomial. Taylor series is widely used in the context of perturbation theory \citep{holmes1998introduction}, where we approximate a function at a perturbed point, $s(x + \epsilon)$, which is also the context of denoising score matching. 

\begin{theorem}
\label{thm:taylo}
    Let $U$ be an open subset of $\mathbb{R}^d$ and $s \in C^{l}(U, \mathbb{R}^d)$ be a $l$-differentiable mapping taking value in $U$ to $\mathbb{R}^d$. For $k \leq l$ and a point $\mathbf{a} \in U$, we define the Taylor polynomial  $T_{s, \mathbf{a}}^k$, using a multi-index notation, such that:
    \begin{equation}
    T_{s, \mathbf{a}}^k(\mathbf{x})=\sum_{|\alpha| \leq k} \frac{(\mathbf{x}-\mathbf{a})^\alpha}{\alpha !} \partial^\alpha s(\mathbf{a}) .
    \end{equation}
    Then the mapping $(\mathbf{a}, x) \rightarrow R_{s, \mathbf{a}}^k = s - T_{s, \mathbf{a}}^k$ is $l-k$ differentiable on $U \times U$ where $R_{s, \mathbf{a}}^k$ is called the remainder. In addition, for every compact $K \subset U$ and every $\delta > 0$ there exists $h > 0$ such that
    \begin{equation}
        \| R_{s, \mathbf{a}}^k(\mathbf{x}) \| \leq \delta \|x -\mathbf{a} \|^k \; \; \text{if} \;\; \mathbf{a}, \mathbf{x} \in K \; \; \text{and} \;\; \|\mathbf{x} -\mathbf{a} \| \leq h
    \end{equation}
\end{theorem}
\paragraph{Remarks} We can re-write the Taylor expansion such that for $\mathbf{x}, \mathbf{z} \in U$ we have 
\begin{equation}
\label{eq:taylor}
    T_{s, \mathbf{x}}^k(\mathbf{x} + \mathbf{z})=\sum_{|\alpha| \leq k} \frac{\mathbf{z}^\alpha}{\alpha !} \partial^\alpha s(\mathbf{x}) .
\end{equation}
Note that in multi-index notations $\mathbf{z}^{\mathbf{\alpha}} = z_1^{\alpha_1} \times \ldots \times z_d^{\alpha_d} \in \mathbb{R}$ , $\partial^{\alpha} = \partial_1^{\alpha_1}\cdots\partial_d^{\alpha_d}$ and $|\alpha| = \alpha_1 + \ldots + \alpha_2$. We sample  $\mathbf{z}$ from $\mathcal{N}(0, I)$ and derive the expectation of the Taylor expansion with respect to  $\mathbf{z}$. For that, we state in \cref{lem:mom} a known result on the moments of a normal distribution \citep{winkelbauer2014moments}, that is, all the odd moments of a normal distribution equal zeros and all the even moments are known in closed form.
\begin{lemma}
\label{lem:mom}
    Let $\mathbf{z}$ be sampled from a standard Gaussian distribution $\mathcal{N}(0, \mathbf{I})$, then all moments equal:
    \begin{equation}
    \begin{aligned}
        &\mathbb{E}\left[\mathbf{z}^{\alpha}\right] = \delta_{\alpha} = 
\begin{cases}
     0 &\text{if} \; |\alpha|=2p+1, \\
      \prod_{i}\omega_{\alpha_i} & \text{if} \; |\alpha| = 2p,\\
\end{cases}
\\ &\text{where,} \;\; \omega_{\alpha_i} = 
\begin{cases}
    0 &\text{if} \; \alpha_i = 2p_i +1 \\
    \frac{(2p_i)!}{{2^p_ip_i!}} &\text{if} \; \alpha_i = 2p_i\\
\end{cases}
    \end{aligned}
    \end{equation}
\end{lemma}
In addition $\mathbb{E}[\mathbf{z}^{\alpha}\mathbf{z}] = (\mathbb{E}[\mathbf{z}^{\alpha}z_1], \ldots, (\mathbb{E}[\mathbf{z}^{\alpha}z_d])^T$ and $\mathbb{E}[\mathbf{z}^{\alpha}z_k] =  \prod_{i}\omega_{\alpha_i + \delta_{ik}}$, where $\delta_{ik}$ the Kronecker delta.
\begin{theorem}
\label{thm:taylorexp}
    Recalling notations from \cref{thm:taylo} and \cref{eq:taylor}, we have 
    \begin{equation}
    \begin{aligned}
        \mathbb{E}_{\mathbf{z}}\left[ T_{s, \mathbf{x}}^k(\mathbf{x} + \mathbf{z}) \right] = \sum_{\substack{|\alpha| \leq k\\ |\alpha| = 2p}} \frac{\delta_{\alpha}}{\alpha !}  \partial^\alpha s(\mathbf{x})
    \end{aligned}
    \end{equation}
\end{theorem}
\Cref{thm:taylorexp} provides a closed-form expectation for any Taylor expansion where the perturbation is sampled from a Gaussian distribution. As this is the case for denoising score matching, we leverage this result to derive a control variate of the training objective.

\subsection{A control variate on the training objective}
\label{sec:cvlsmall}
We recall the training objective $L_{\boldsymbol{\theta}}(\mathbf{z}, \mathbf{x}, \sigma) =  \frac{1}{2}\left\| \frac{\mathbf{z}}{\sigma} + s_{\boldsymbol{\theta}}(\mathbf{x} + \sigma \mathbf{z})\right\|^2$ and approximate the score network with a Taylor expansion of order $k$ around the data point $\mathbf{x}$. We derive the approximation  and the control variate in \cref{app:cvl_small}, and provide the result here:
\begin{equation}
\label{eq:cvlk}
\begin{aligned}
    &C^{k}_{\boldsymbol{\theta}}(\mathbf{z}, \mathbf{x}, \sigma) =\, \frac{\|\mathbf{z}\|^2 -D}{2 \sigma^2}   \\ 
    &+\frac{1}{2} \sum_{\substack{|\alpha| \leq k \\|\rho| \leq k}}  \frac{\sigma^{|\boldsymbol{\alpha}| + |\boldsymbol{\rho}|}}{\alpha !\rho !} \left(\mathbf{z}^{\boldsymbol{\alpha} + \boldsymbol{\rho}} - \delta_{\boldsymbol{\alpha} + \boldsymbol{\rho}}\right)  \partial^\alpha s_{\boldsymbol{\theta}}(\mathbf{x})^T \partial^\rho s_{\boldsymbol{\theta}}(\mathbf{x}) \\
    &+ \sum_{|\alpha| \leq k}  \frac{\sigma^{|\boldsymbol{\alpha}|-1}}{ \alpha !} \left(\mathbf{z}^{\boldsymbol{\alpha}}\mathbf{z}^T -\mathbb{E}[\mathbf{z}^{\alpha}\mathbf{z}] \right) \partial^\alpha  s_{\boldsymbol{\theta}}(\mathbf{x}) .
\end{aligned}
\end{equation}
Note that the control variate has \textit{zero expectation} with respect to $\mathbf{z}$, by applying \cref{thm:taylorexp} and \cref{lem:mom}. In addition, we introduce a regression coefficient $\beta$ and the training objective $ L_{\boldsymbol{\theta}, \beta}^{k}$, given by 
\begin{equation}
\label{eq:loss_cvl}
    L_{\boldsymbol{\theta}, \beta}^{\text{cvl}, k}(\mathbf{z}, \mathbf{x},  \sigma) = \lambda(\sigma)\left(L_{\boldsymbol{\theta}}(\mathbf{z}, \mathbf{x}, \sigma) - \beta C_{\boldsymbol{\theta}}^k(\mathbf{z}, \mathbf{x}, \sigma)\right) .
\end{equation}
For any $\beta$, $ L_{\boldsymbol{\theta}, \beta}^{k}$ equals $L$, so the training objective in expectation is maintained and only its variance is affected. As we will show in the experiments, $\beta$ greatly influences the reduction in variance, and it is key to set it as close as possible to the optimal value $\beta_{\text{opt}}$.
Note that this control variate is a generalisation of the one derived by \cite{wang2020wasserstein}, which we obtain by setting $k=0$ and $\beta=1$. In \cref{sec:cv12}, we derive the control variate for $k=1$ and $k=2$.

\subsection{A control variate on the gradients}

\cite{NIPS2013_9766527f} shows that excessive variance in the gradient estimator leads to longer convergence and thus argues for reducing the variance of the gradients. Following that line of thought, we derive a control variate on the gradient of the training objective, $\partial_{\boldsymbol{\theta}}L_{\boldsymbol{\theta}}(\mathbf{z}, \mathbf{x},  \sigma)$,  using the same methodology as in \cref{sec:cvlsmall}. We begin by deriving the gradient and approximate the score $s_{\boldsymbol{\theta}}(\mathbf{x} + \sigma \mathbf{z})$ and its gradient $\partial_{\boldsymbol{\theta}}s_{\boldsymbol{\theta}}(\mathbf{x} + \sigma \mathbf{z})$ with Taylor expansions. We derive the approximation and the control variate in \cref{app:cvg_small} and obtain the control variate for the gradient 
\begin{equation}
\begin{aligned}
    &C_{\textbf{g}, \boldsymbol{\theta}}^{k}(\mathbf{z}, \mathbf{x}, \sigma) =\, \sum_{|\rho| \leq k}  \frac{\sigma^{|\boldsymbol{\rho}|-1}}{ \rho !} \left(\mathbf{z}^{\boldsymbol{\rho}} \mathbf{z} - \mathbb{E}[\mathbf{z}^{\rho}\mathbf{z}]\right)^T\partial^\rho  \partial_{\boldsymbol{\theta}}s_{\boldsymbol{\theta}}(\mathbf{x}) \\
    & + \sum_{\substack{|\rho| \leq k \\ |\alpha| \leq k}} \frac{\sigma^{|\boldsymbol{\alpha}| + |\boldsymbol{\rho}|}}{\alpha! \rho!} \left(\mathbf{z}^{\boldsymbol{\alpha} + \boldsymbol{\rho}} - \delta_{\boldsymbol{\alpha} + \boldsymbol{\rho}}\right)\partial^\alpha  s_{\boldsymbol{\theta}}(\mathbf{x})^T \partial^\rho  \partial_{\boldsymbol{\theta}}s_{\boldsymbol{\theta}}(\mathbf{x}).
\end{aligned}
\end{equation}
Note that each parameter of the network is individually controlled. We introduce a regression coefficient $\beta_{\mathbf{g}}$ to scale the control variate for each parameter. If we set $k = 0$, we recover the gradient of the objective's control variate derived by \cite{wang2020wasserstein}. This hints at an equivalence between the control variate on the training objective and on the gradients, which we prove in the following section

\subsection{Controlling the training objective is equivalent to controlling its gradient}
\label{sec:proof_small}
The previous result suggests an equivalence between the control variate of the objective and of the gradients. Indeed, \cref{thm:eqvalence} proves this claim to any $k$-th order Taylor approximation. We prove (\cref{app:proof_eqvalence}) that controlling the training objective is equivalent to controlling its gradients. We derive the gradients of the objective's control variate, $\partial_{\boldsymbol{\theta}}C^{k}_{\boldsymbol{\theta}}(\mathbf{z}, \mathbf{x}, \sigma) $ and prove that it equals the gradient's control variate $\mathcal{C}^{k, k}_{\boldsymbol{\theta}}(\mathbf{z}, \mathbf{x}, \sigma) $:
\begin{theorem}
\label{thm:eqvalence}
Let $C^{k}_{\boldsymbol{\theta}}(\mathbf{z}, \mathbf{x}, \sigma)$ be the control variate on the training objective and $C^{k}_{\mathbf{g}, \boldsymbol{\theta}}(\mathbf{z}, \mathbf{x}, \sigma)$ the control variate on the training objective's gradient, we have the equality:
    \begin{align}
 \partial_{\boldsymbol{\theta}}C^{k}_{\boldsymbol{\theta}}(\mathbf{z}, \mathbf{x}, \sigma) = C^{k}_{\mathbf{g}, \boldsymbol{\theta}}(\mathbf{z}, \mathbf{x}, \sigma)
    \end{align}
\end{theorem}
This equality explains the benefits observed by \cite{wang2020wasserstein}. However, the regression coefficient of the objective's control variate is unrelated to that of the gradients. Since this coefficient is decisive for the quality of the control variate, we cannot expect to control the variance of the gradient through the objective alone. That comes as an unfortunate cost. Indeed, computing the regression coefficient $\beta$ for the training objective is inexpensive since it only involves computing a batch of training loss values. Conversely, computing the regression coefficient $\beta_{\mathbf{g}}$ for the gradients is expensive, as it requires the gradients in batches, which is memory-intensive. In addition, we require a reliable estimate of  $\beta_{\mathbf{g}}$, which necessitates a large batch size. 
\subsection{A control variate for large values of $\sigma$}

We derived the previous control variate around the data point $\mathbf{x}$ and considered $\sigma \mathbf{z}$ to be the perturbation. While that approach is valid for small values of $\sigma$, in score-based modelling, $\sigma$ ranges up to $100$ \citep{song2021scorebased}.  In such cases, this assumption does not hold, and the Taylor expansion is of poor quality, negatively impacting the training \citep{song2021train}. When $\|\sigma \mathbf{z}\| \geq \|\mathbf{x}\|$, it is more appropriate to derive the Taylor series around $\sigma \mathbf{z}$ and consider  $\mathbf{x}$ to be the perturbation. 
%
%
%
We present the control variate on the training objective
\begin{equation}
\begin{aligned}
    &\mathcal{C}^{k}_{\boldsymbol{\theta}}(\mathbf{z}, \mathbf{x}, \sigma) =\, \sum_{|\alpha| \leq k}  \frac{1}{ \sigma\alpha !} \left(\mathbf{x}^{\boldsymbol{\alpha}}  - \boldsymbol{\mu}_{|\boldsymbol{\alpha}|}\right)  \mathbf{z}^T \partial^\alpha  s_{\boldsymbol{\theta}}(\sigma\mathbf{z}) \\
    & +\frac{1}{2} \sum_{\substack{|\alpha| \leq k \\|\rho| \leq k}}  \frac{1}{\alpha !\rho !} \left(\mathbf{x}^{\boldsymbol{\alpha} + \boldsymbol{\rho}} - \boldsymbol{\mu}_{|\boldsymbol{\alpha}| + |\boldsymbol{\rho}|}\right) \partial^\alpha s_{\boldsymbol{\theta}}(\sigma\mathbf{z})^T \partial^\rho s_{\boldsymbol{\theta}}(\sigma\mathbf{z}) 
\end{aligned}
\end{equation}
and on the gradients of the training objective
\begin{equation}
\begin{aligned}
    &\mathcal{C}_{\textbf{g}, \boldsymbol{\theta}}^{k}(\mathbf{z}, \mathbf{x}, \sigma) =\, \sum_{|\rho| \leq k}  \frac{1}{\sigma \rho !} \left(\mathbf{x}^{\boldsymbol{\rho}} - \mu_{|\boldsymbol{\rho}|}\right)  \mathbf{z}^T\partial^\rho  \partial_{\boldsymbol{\theta}}s_{\boldsymbol{\theta}}(\sigma\mathbf{z})  \\
    &+   \sum_{\substack{|\alpha| \leq k \\|\rho| \leq k}} \frac{1}{\alpha! \rho!} \left(\mathbf{x}^{\boldsymbol{\alpha} + \boldsymbol{\rho}} - \mu_{|\boldsymbol{\alpha}| + |\boldsymbol{\rho}|}\right)  \partial^\alpha  s_{\boldsymbol{\theta}}(\sigma\mathbf{z})^T \partial^\rho  \partial_{\boldsymbol{\theta}}s_{\boldsymbol{\theta}}(\sigma\mathbf{z}) 
\end{aligned}
\end{equation}
and refer to \cref{app:cvg_large} for their derivation. Additionally, we note $\mu_n$ the $n$-th moment of the data, $\mu_n = \mathbb{E}[\mathbf{x}^n]$.
In our experiments, we use $k=1$ and normalise the data such that $\mathbf{\mu}_1=\mathbf{0}$.
Lastly, a similar derivation as the one done in \cref{sec:proof_small} shows that the gradients with respect to the parameter of $\mathcal{C}_{\boldsymbol{\theta}}^{k}$ equals the control variate on the gradients, $\partial_{\boldsymbol{\theta}}\mathcal{C}^{k}_{\boldsymbol{\theta}}(\mathbf{z}, \mathbf{x}, \sigma) = \mathcal{C}_{\textbf{g}, \boldsymbol{\theta}}^{k}(\mathbf{z}, \mathbf{x}, \sigma)$. This shows that it is equivalent to controlling the training objective or its gradients.

\section{Experiments}

We are now equipped with two sets of control variate, $(C^{k}_{\boldsymbol{\theta}}, C_{\textbf{g}, \boldsymbol{\theta}}^{k})$ and $(\mathcal{C}^{k}_{\boldsymbol{\theta}}, \mathcal{C}_{\textbf{g}, \boldsymbol{\theta}}^{k})$, and we study their ability to reduce variance. We measure the reduction in variance of the training objective as the ratio $\rho_{\beta} = \operatorname{Var}(L_{\boldsymbol{\theta}} - \beta C_{\boldsymbol{\theta}}^k) / \operatorname{Var}(L_{\boldsymbol{\theta}})$, and similarly we measure the reduction in the variance of the gradients and denotes it $\rho_{\beta,\mathbf{g}}$. A ratio smaller than one indicates a reduction in variance, and lower is better.

We consider three sets of experiments: (1) we explore variance reduction on a toy dataset and show a setup where control variate enables convergence to the solution; (2) we explore variance reduction on MNIST and show the limitation of control variate in this setting; (3) we study the variance reduction on an MLP of varying width and depth and justify the limitation.
\subsection{Control variate on a toy dataset}
\label{sec:toydatasec}
In the following set of experiments, we train an MLP on a two-dimensional, bi-modal, Gaussian distribution generated by $p(x) = 1/5 \mathcal{N}(x; \mathbf{5}, \mathbf{I}) +  4/5 \mathcal{N}(x; -\mathbf{5}, \mathbf{I})$. We (1) will show the necessity to control the gradients with a regression coefficient $\beta$; and (2) reduce the variance of the gradients for $\sigma \in [0.1, 90]$ using $ C_{\textbf{g}, \boldsymbol{\theta}}^{1}$ and $\mathcal{C}_{\textbf{g}, \boldsymbol{\theta}}^{1}$; (3) present a setup where control variate enables the convergence to the solution; (4) compare $C_{\textbf{g}, \boldsymbol{\theta}}^{0}$, $C_{\textbf{g}, \boldsymbol{\theta}}^{1}$ and $C_{\textbf{g}, \boldsymbol{\theta}}^{2}$.

\subsubsection{$\beta$'s impact}
\label{subsec:beta}
We will now demonstrate the importance of $\beta$ and underscore the need to control the gradients, not the training objective. We control the variance of the training objective with $C^{1}_{\boldsymbol{\theta}}$ with and without the regression coefficient $\beta$. In addition, we also measure the reduction in the variance of the gradients, \textit{while controlling the objective only}. \Cref{tab:beta_study} reports that using $\beta$ is always beneficial, and not using it drastically \textit{increases} variance. \Cref{tab:grad_var_of_cvl} reports that even though there is an equivalence between controlling the objective and the gradients, the variance of the gradients increases, regardless of the use of $\beta$. This increase comes from the regression coefficient $\beta$ being designed specifically for the objective rather than its gradients. This evidence supports our argument of the necessity to control the gradients and not the objective.

\begin{table}[h] 
    \centering
    \setlength\tabcolsep{3pt}
    \caption{Reduction in the variance of the training objective when using control variate, for the toy dataset. Without a regression coefficient, the variances greatly increase for $\sigma > 1$. This confirms the behaviour reported by \cite{song2021train}. On the contrary, using $\beta$ always reduces the variance regardless of $\sigma$.}
    \vspace{-5pt} 
    \resizebox{\columnwidth}{!}{ 
        \begin{tabular}{lccccc}
            \toprule
            $\sigma$ & 0.1 & 0.5 & 1 & 5 & 10 \\ \midrule
            $\rho_{\beta}$ & 0.00\tiny{$\pm 0.00$} & 0.05\tiny{$\pm0.01$} & 0.33\tiny{$\pm0.06$} & 0.81\tiny{$\pm0.08$} & 0.82\tiny{$\pm0.07$} \\ 
            $\rho_{\beta=1}$ & 0.00\tiny{$\pm0.00$} & 0.25\tiny{$\pm0.04$} & 2.92\tiny{$\pm0.57$} & 3.25\tiny{$\pm1.31$} & 24.17\tiny{$\pm5.95$} \\ 
            \bottomrule
        \end{tabular}
    }
    \label{tab:beta_study}
\end{table}
\begin{table}[h] 
    \caption{Reduction in the variance of the training objective's gradient, for the toy dataset.}
    \vspace{-5pt} 
    \resizebox{\columnwidth}{!}{
        \begin{tabular}{lccccc}
            \toprule
            $\sigma$ & 0.1 & 0.5 & 1 & 5 & 10 \\ \midrule
            $\rho_{\mathbf{g}, \beta}$ & 0.01\tiny{$\pm 0.00$} & 0.27\tiny{$\pm0.03$} & 0.63\tiny{$\pm0.05$} & 1.25\tiny{$\pm0.18$} & 1.17\tiny{$\pm0.10$} \\ 
            $\rho_{\mathbf{g}, \beta=1}$ & 0.01\tiny{$\pm0.00$} & 0.31\tiny{$\pm0.04$} & 1.09\tiny{$\pm0.13$} & 7.07\tiny{$\pm1.11$} & 25.77\tiny{$\pm6.74$} \\
            \bottomrule
        \end{tabular}
    }
    \label{tab:grad_var_of_cvl}
\end{table}

\subsubsection{Controlling the objective's gradients}

We will now proceed to control the objective's gradients with the control variate $C_{\textbf{g}, \boldsymbol{\theta}}^{1}$, and $\mathcal{C}_{\textbf{g}, \boldsymbol{\theta}}^{1}$ and report our results in \cref{tab:var_red_g}. This experiment shows that either the former or the latter control variate \textit{reduces the variance of the gradients for any $\sigma$ value in the range $[0.1, 90]$}, showing the effectiveness of our approach in that problem setting. 

\begin{table*}[]
\caption{Controlling the objective's gradient with a regression coefficient $\beta_{\mathbf{g}}$ always reduces the variance, at worst, has no impact. The first row indicates the variance reduction across $\sigma$ when using the control variate designed for small values of $\sigma$. For $\sigma<1$ there is a substantial decrease in variance, while for $\sigma > 1$, $\beta_{\mathbf{g}}$ ensures no increase in variance. The third row shows variance reduction when using control variate designed for large values of sigma. As expected the reduction is limited for $\sigma < 5$, but for $\sigma > 5$ more than half of the variance is reduced.}
\setlength\tabcolsep{3.0pt}
\resizebox{\textwidth}{!}{
\begin{tabular}{ccccccccccc}
\toprule
$\sigma$ & 0.1 & 0.5 & 1 & 5 & 10 & 20 & 40 & 60 & 80 & 90 \\ \midrule
$\rho_{\mathbf{g}}$ small & $0.02$\tiny{$\pm 0.00$} & $0.25$\tiny{$\pm 0.03$} & $0.56$\tiny{$\pm 0.04$} & $0.82$\tiny{$\pm 0.02$} & $0.84$\tiny{$\pm 0.02$} & $0.86$\tiny{$\pm 0.02$} & $0.89$\tiny{$\pm 0.02$} & $0.91$\tiny{$\pm 0.02$} & $0.93$\tiny{$\pm 0.02$} & $0.94$\tiny{$\pm 0.02$} \\
$\beta_{\mathbf{g}}$ & $0.99$\tiny{$\pm 0.01$} & $0.98$\tiny{$\pm 0.02$} & $0.98$\tiny{$\pm 0.04$} & $0.85$\tiny{$\pm 0.06$} & $0.49$\tiny{$\pm 0.08$} & $0.39$\tiny{$\pm 0.07$} & $0.40$\tiny{$\pm 0.12$} & $0.41$\tiny{$\pm 0.21$} & $0.34$\tiny{$\pm 0.21$} & $0.35$\tiny{$\pm 0.31$} \\
$\rho_{\mathbf{g}}$ large & $0.70$\tiny{$\pm 0.04$} & $0.87$\tiny{$\pm 0.04$} & $0.91$\tiny{$\pm 0.03$} & $0.80$\tiny{$\pm 0.02$} & $0.47$\tiny{$\pm 0.04$} & $0.26$\tiny{$\pm 0.05$} & $0.30$\tiny{$\pm 0.10$} & $0.49$\tiny{$\pm 0.13$} & $0.60$\tiny{$\pm 0.13$} & $0.59$\tiny{$\pm 0.14$} \\
$\beta_{\mathbf{g}}$ & $0.34$\tiny{$\pm 0.16$} & $0.19$\tiny{$\pm 0.09$} & $0.14$\tiny{$\pm 0.06$} & $0.34$\tiny{$\pm 0.05$} & $0.79$\tiny{$\pm 0.05$} & $0.85$\tiny{$\pm 0.02$} & $0.82$\tiny{$\pm 0.02$} & $0.79$\tiny{$\pm 0.04$} & $0.73$\tiny{$\pm 0.06$} & $0.72$\tiny{$\pm 0.06$} \\ 
\bottomrule
\end{tabular}
}
\label{tab:var_red_g}
\end{table*}

\subsubsection{Enabling convergence}
\label{sec:fast_cvg}
In this experiment we control the gradients with $C_{\textbf{g}, \boldsymbol{\theta}}^{2}$ for $\sigma \in [0.01, 1]$ and set the batch size to equal ten. We observe that the setup without control variate does not converge, whereas the setup with control variate does. However, if we increase the batch size to 1000, then both setups converge. This indicates that in a small batch size regime, where we expect high variance, and with good Taylor approximation, the control variate enables the convergence to the solution.
\begin{figure}[h] 
    \centering
    \begin{subfigure}{}
        \includegraphics[width=0.47\columnwidth]{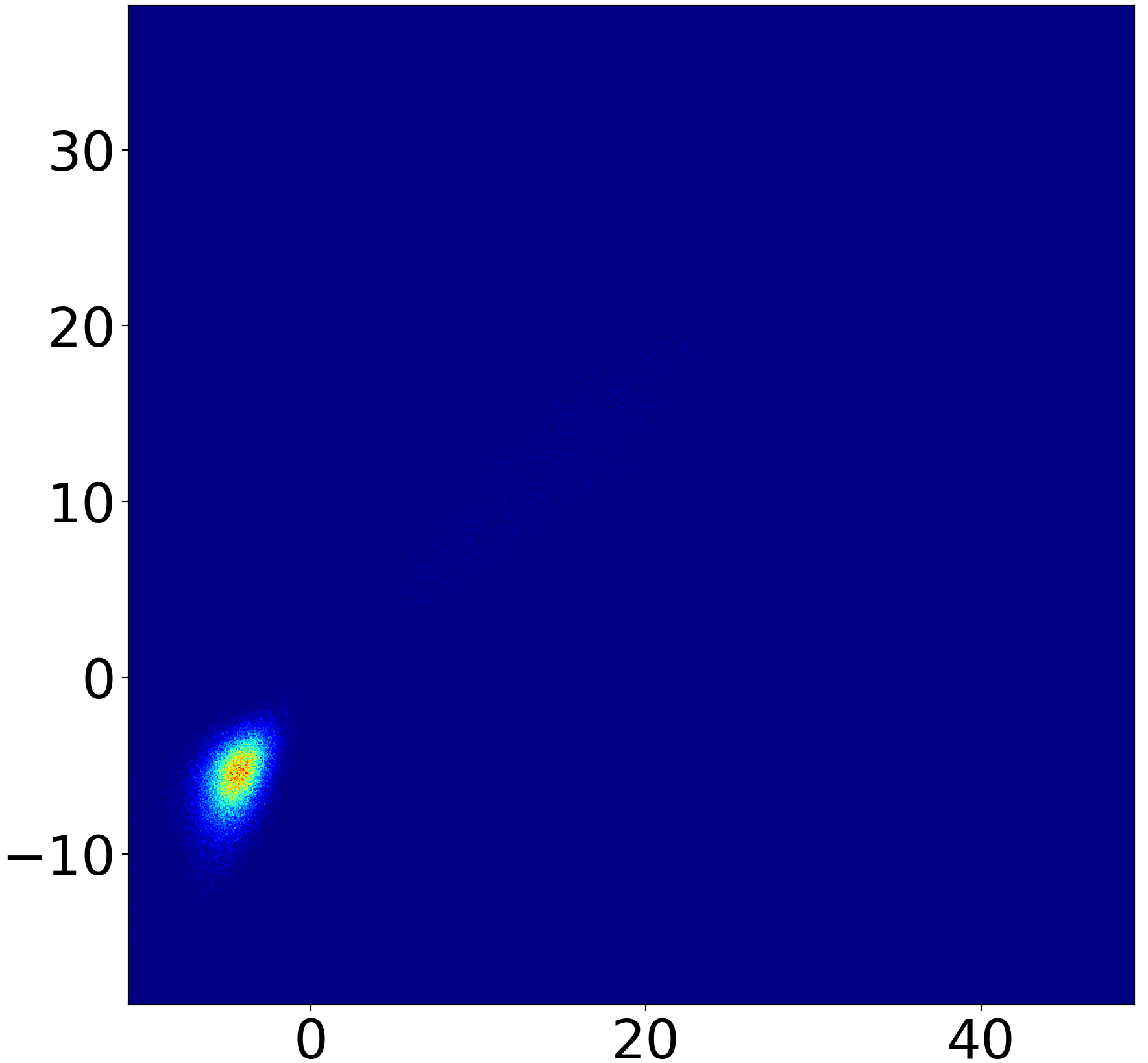}
        \label{fig:figure1}
    \end{subfigure}
    \hfill
    \begin{subfigure}{}
        \includegraphics[width=0.47\columnwidth]{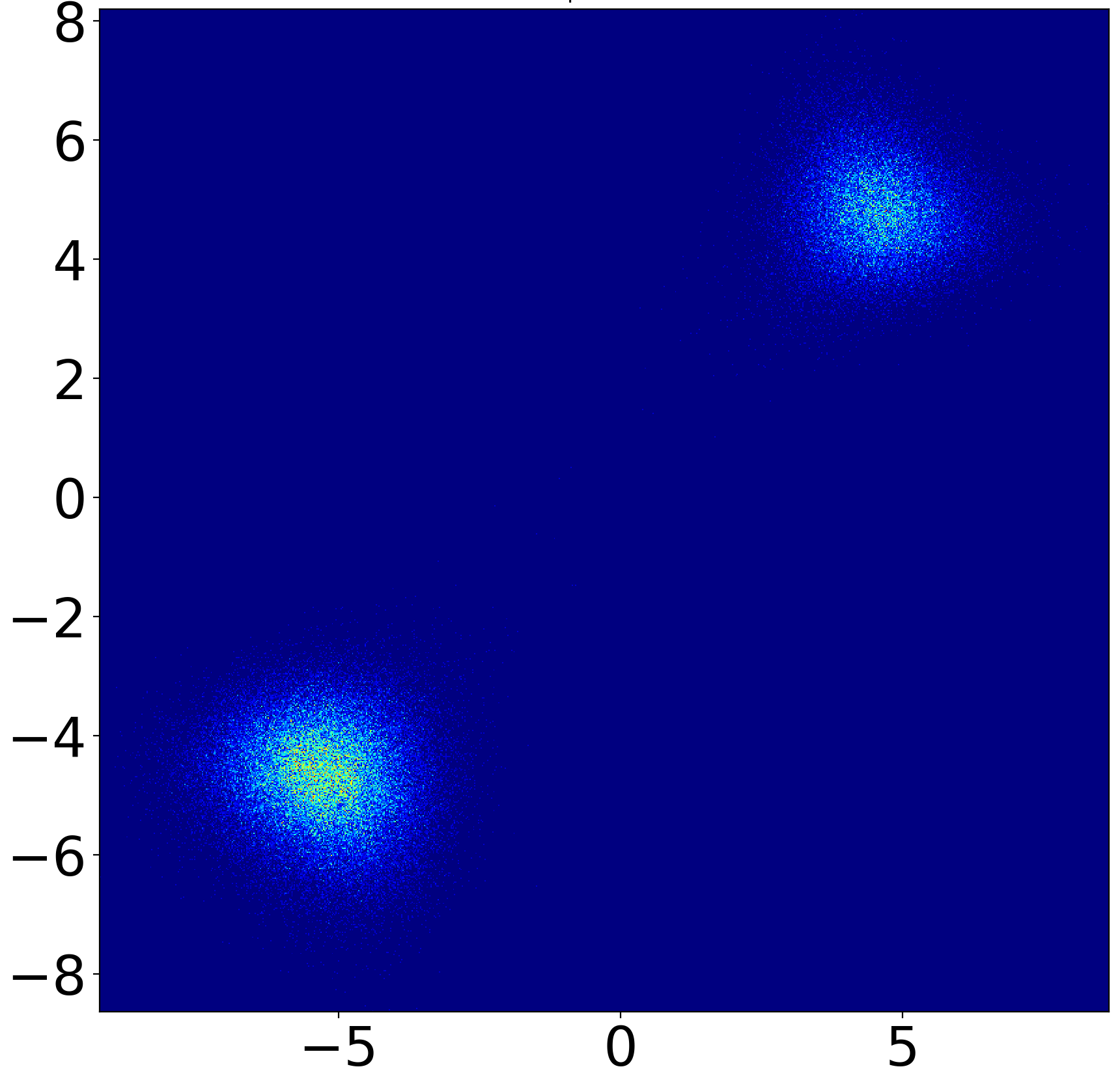}
        \label{fig:figure2}
    \end{subfigure}
    \vspace{-1em} 
    \caption{Convergence with (right) and without control variate (left)}
    \label{fig:sidebyside}
\end{figure}

\subsubsection{Comparing $C_{\textbf{g}, \boldsymbol{\theta}}^{0}$, $C_{\textbf{g}, \boldsymbol{\theta}}^{1}$ and $C_{\textbf{g}, \boldsymbol{\theta}}^{2}$}

\begin{figure}  
    \caption{Variance reduction (right) and regression coefficient (left) for $C_{\textbf{g}, \boldsymbol{\theta}}^{0}$, $C_{\textbf{g}, \boldsymbol{\theta}}^{1}$ and $C_{\textbf{g}, \boldsymbol{\theta}}^{2}$}
    \centering
    \begin{subfigure}{} 
        \includegraphics[width=0.47\columnwidth]{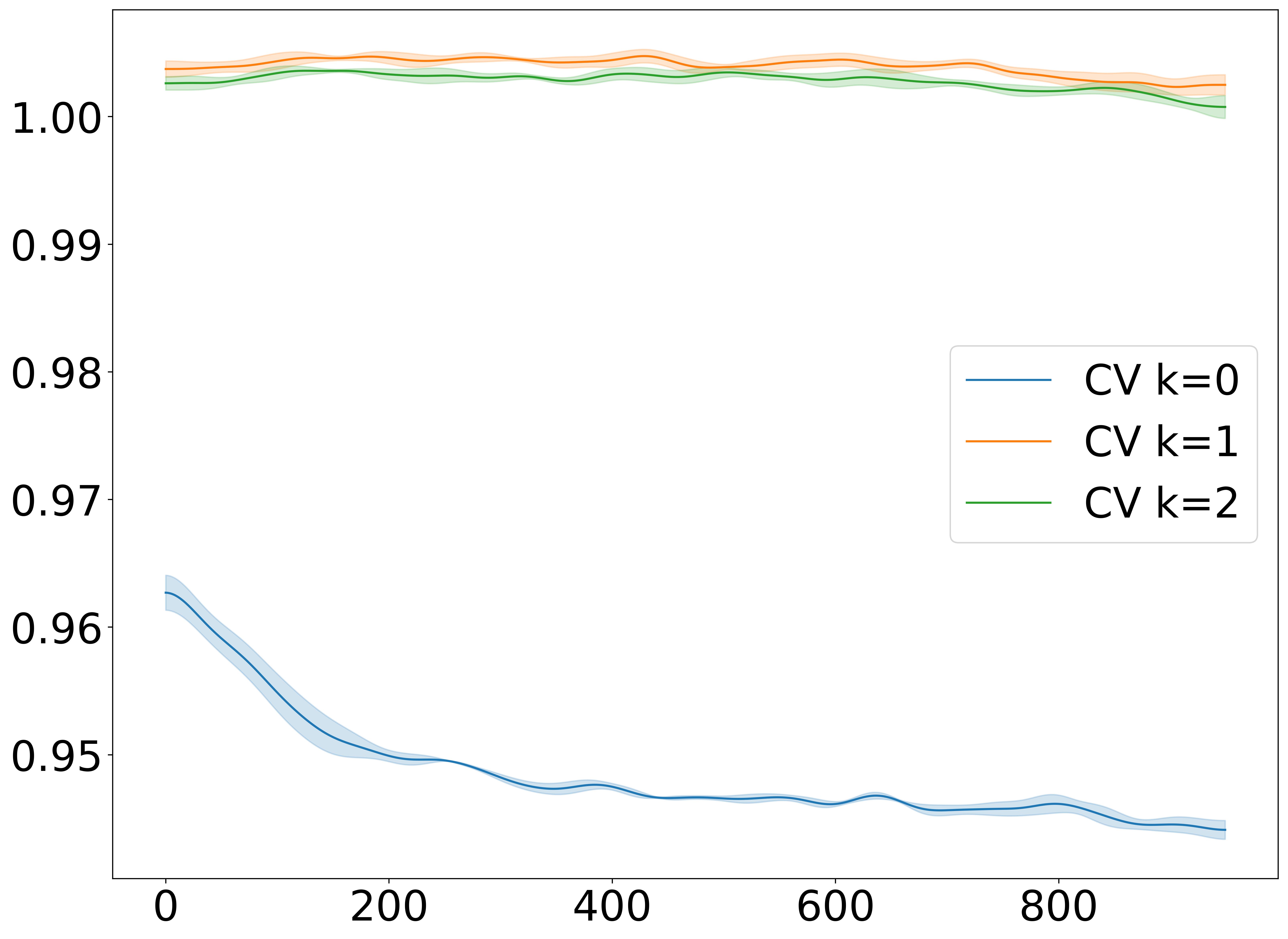}
        \label{fig:betacvg012}
    \end{subfigure}
    \hfill
    \begin{subfigure}{}
        \includegraphics[width=0.47\columnwidth]{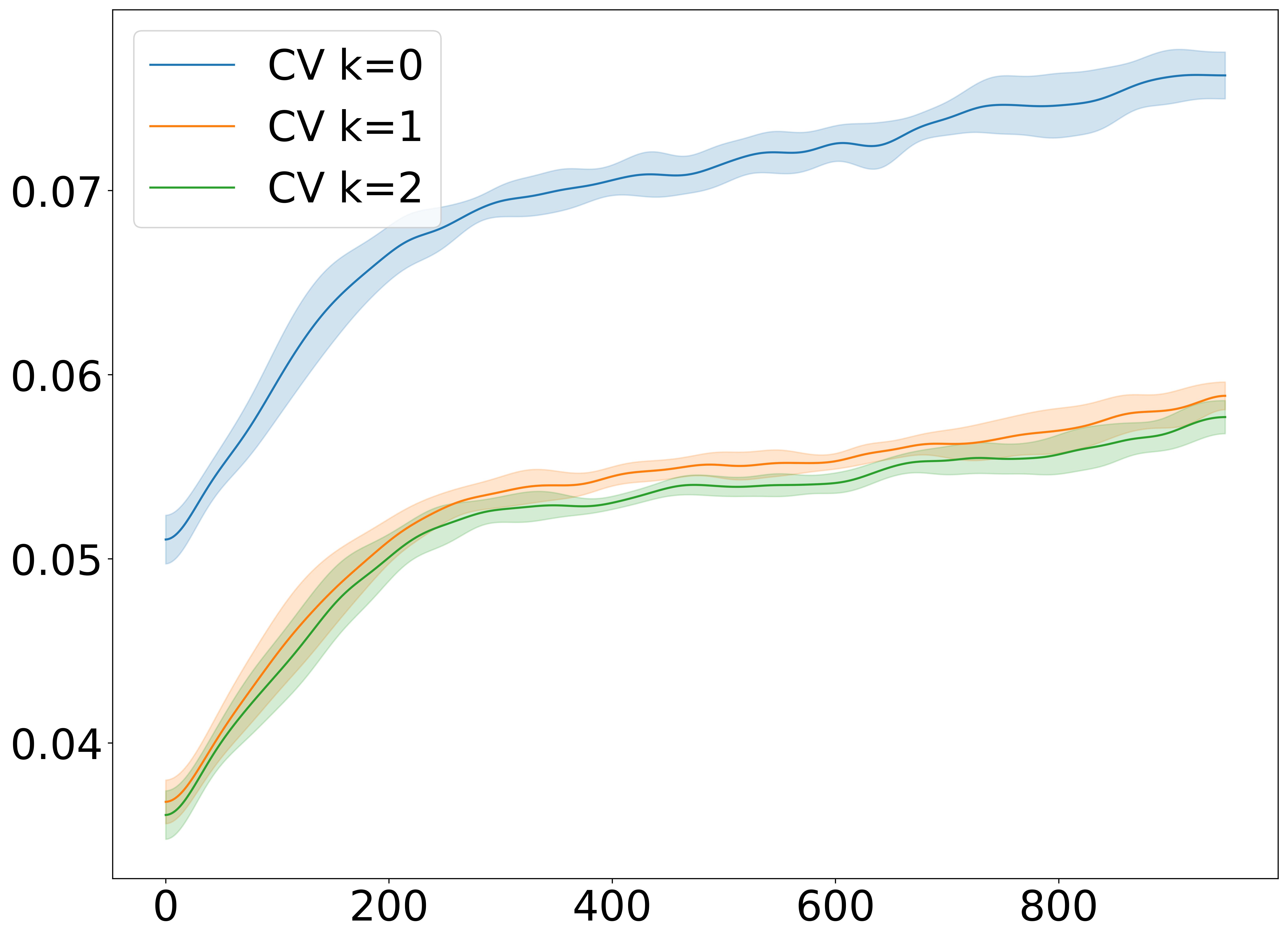}
        \label{fig:varcvg012}
    \end{subfigure}
    \vspace{-1em} 
    \label{fig:sidebyside}
\end{figure}

We will now compare the three control variate $C_{\textbf{g}, \boldsymbol{\theta}}^{0}$, $C_{\textbf{g}, \boldsymbol{\theta}}^{1}$ and $C_{\textbf{g}, \boldsymbol{\theta}}^{2}$. We observe an improvement in the variance reduction between $k=0$ and $k=1$ and a marginal one between $k=1$ and $k=2$, \cref{fig:varcvg012}. This would suggest that the MLP behaves loosely as a linear function.

\subsection{Variance reduction on MNIST}
\label{sec:varredmnist}
\begin{figure}  
    \caption{Variance reduction (right) and training loss (left) on MNIST}
    \centering
    \begin{subfigure}{}
        \includegraphics[width=0.47\linewidth]{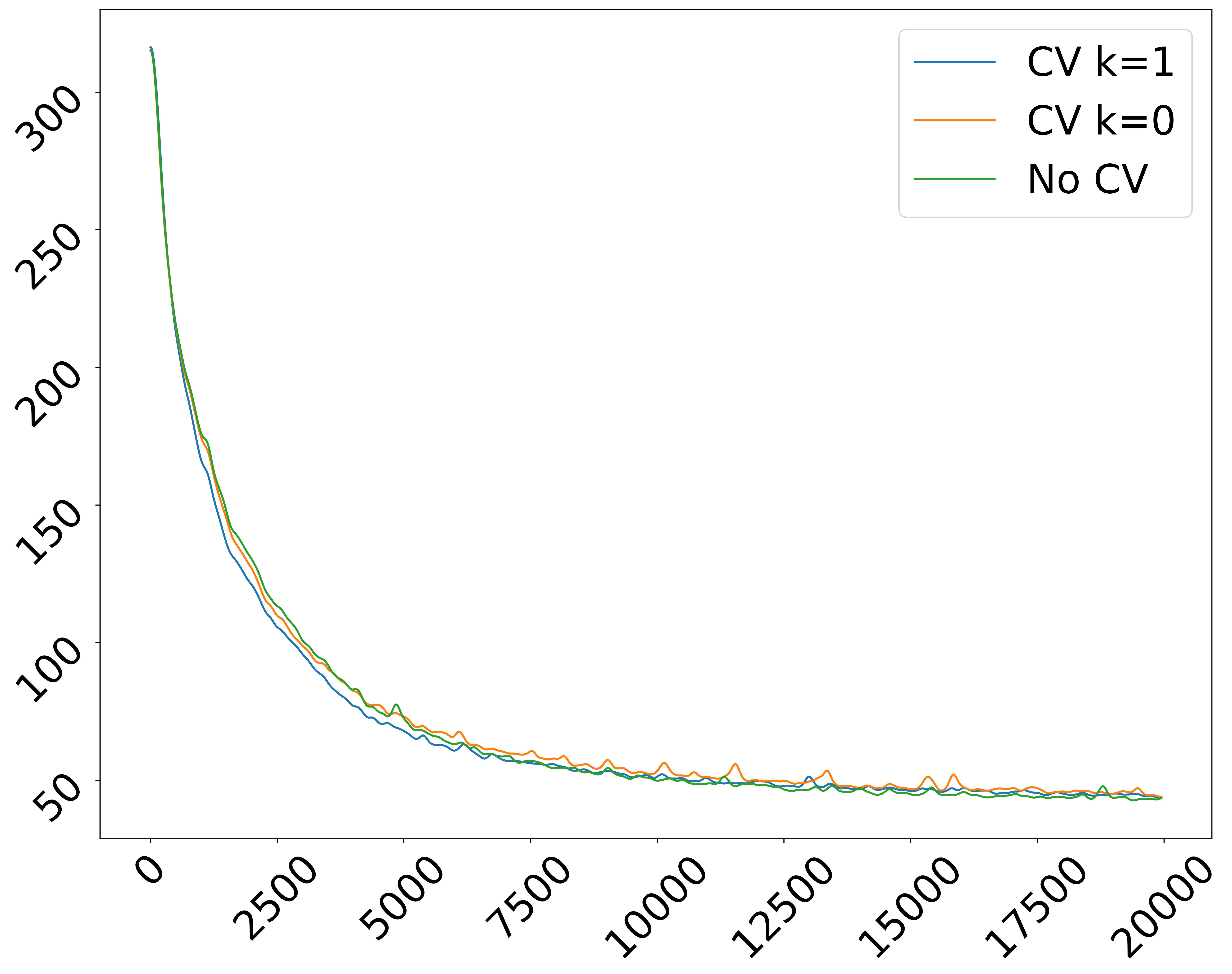}
        \label{fig:loss_mnist}
    \end{subfigure}
    \hfill
    \begin{subfigure}{}
        \includegraphics[width=0.47\linewidth]{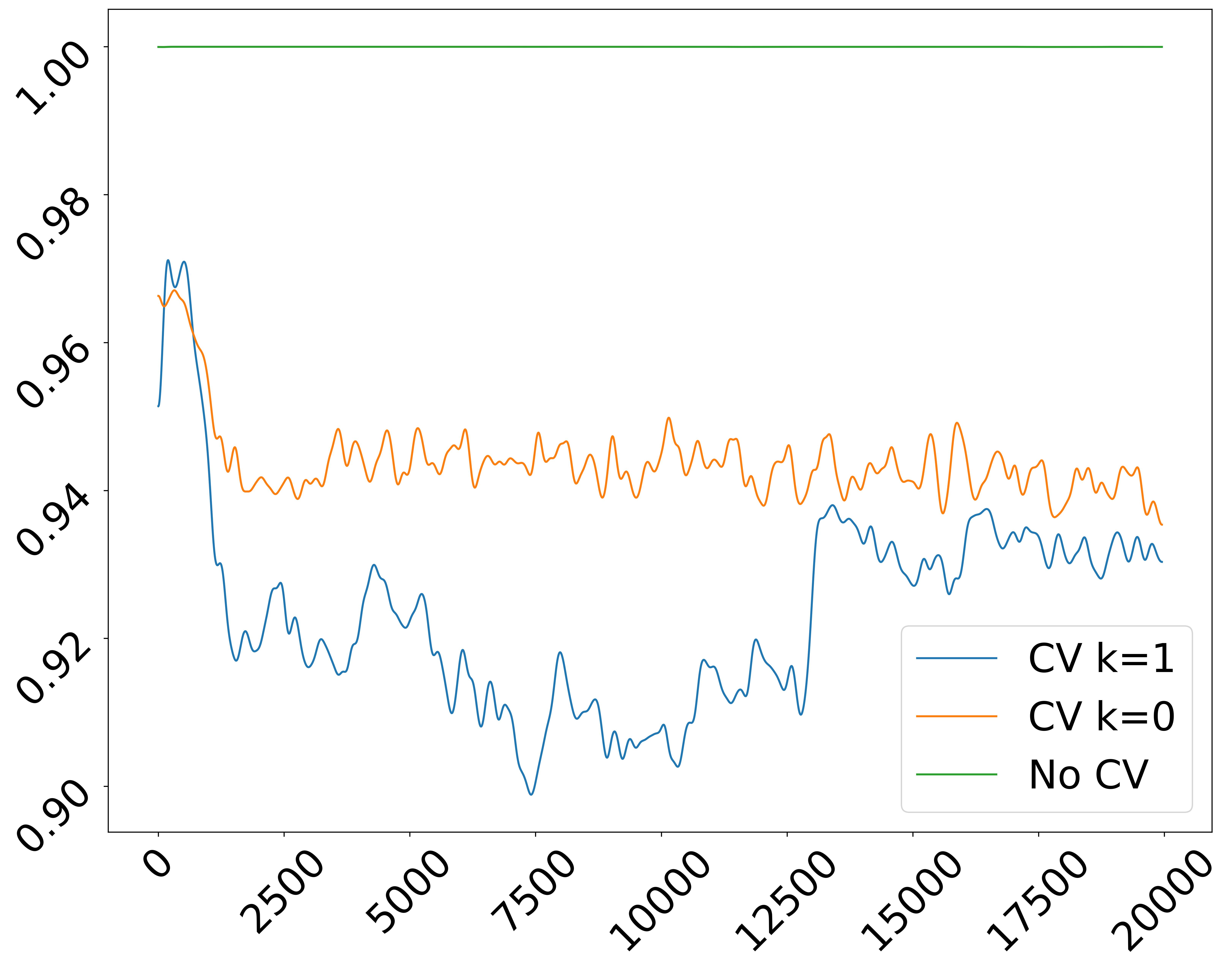}
        \label{fig:varred_mnist}
    \end{subfigure}
    \vspace{-1em} 
    \vspace{-2\baselineskip} 
\end{figure}

In the previous experiments, we have shown in a simple setting the benefit of using the control variate, when variance is high. We will now proceed with a set of experiments on a more challenging dataset and complex model. In this experiment we control the gradients of a U-Net, training on MNIST. We experiment with $C_{\textbf{g}, \boldsymbol{\theta}}^{1}$ and $C_{\textbf{g}, \boldsymbol{\theta}}^{0}$. We observe a better variance reduction when using $k=1$ than $k=0$ (\cref{fig:varred_mnist}). However, the variance reduction is marginal and yields no benefit on the convergence of the loss (\cref{fig:loss_mnist}). We suppose three explanations: (1) the variance reduction is not beneficial and is something that might not need to be addressed; (2) a Taylor approximation with $k=1$ and $k=2$ is a poor approximation for large neural networks such as a U-Net. We investigate further this hypothesis in \cref{sec:irregularity}; (3) the optimiser used, Adam, already has a variance reduction mechanism and deals with its most harmful effect.
\begin{figure}  
    \caption{Variance reduction on toy dataset comparing Adam and SGD}
    \centering
    \begin{subfigure}{}
        \includegraphics[width=0.8\columnwidth]{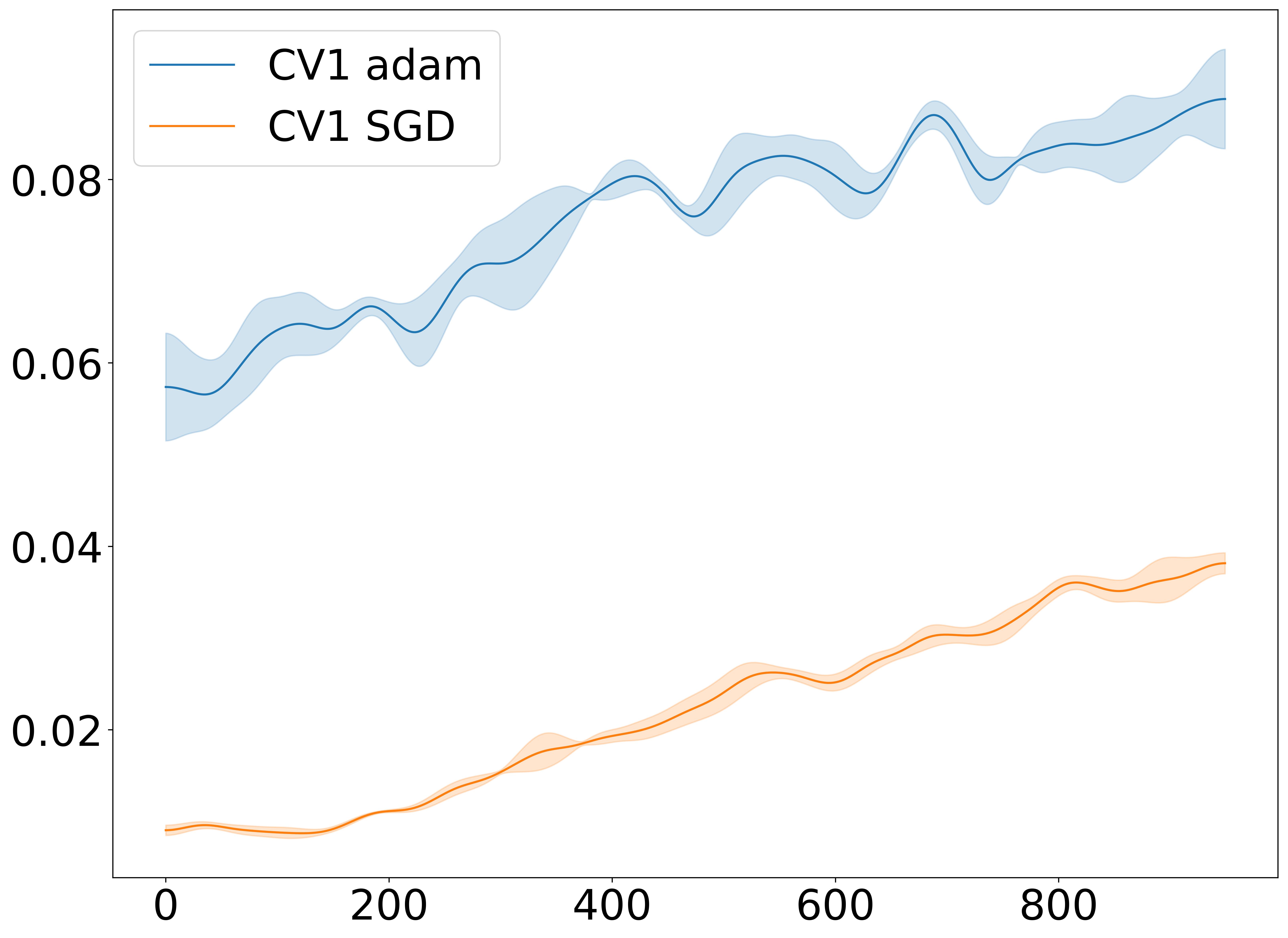}
        \label{fig:varred_mlp_sgdadam}
    \end{subfigure}
    \vspace{-1em} 
\end{figure}

\subsection{Studying the variance reduction with respect to the irregularity of the network}
\label{sec:irregularity}

The previous experiment showed poor variance reduction, one possible explanation is that the Taylor expansion poorly approximates the gradients of complex maps, such as a U-Net. However, the variance reduction was significant when using a small MLP, suggesting a faithful Taylor approximation. Thus, we hypothesise that the approximation quality decreases with increased network capacity, which we attempt to confirm in the following experiment.

\begin{figure}[h]
    \centering
    \begin{minipage}{0.95\columnwidth}
        \caption{Average variance reduction for various MLP configurations (lower is better). In each box the value is the variance reduction and in parenthesis is the number of parameters.}
        \centering
        \includegraphics[width=\columnwidth]{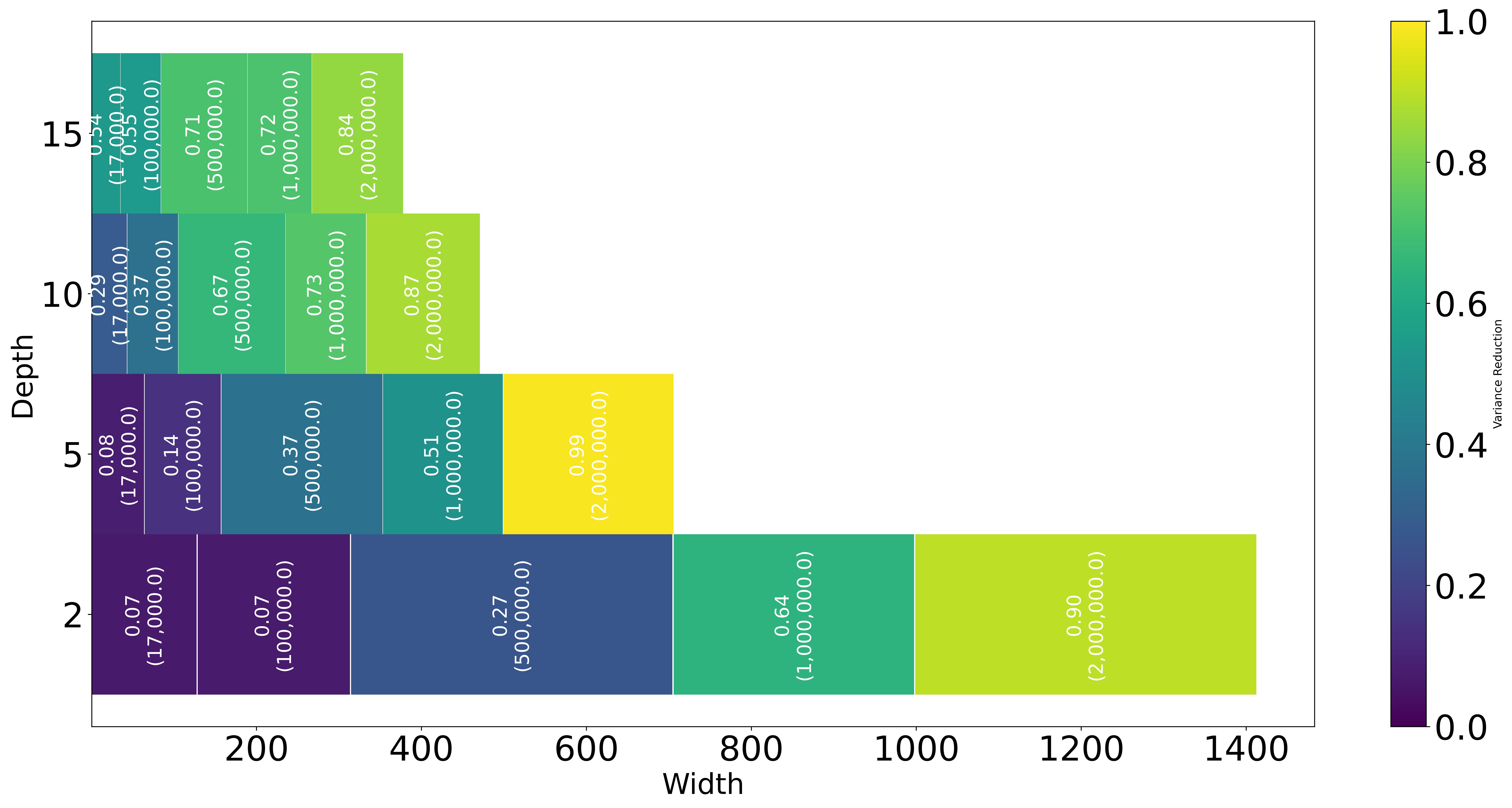} 
        \label{fig:width_depth_varred}
    \end{minipage}\hfill
    \begin{minipage}{0.95\columnwidth}
        \caption{Average variance reduction for various MLP configurations with spectral normalisation (lower is better). In each box the value is the variance reduction and in parenthesis is the number of parameters.}
        \centering
        \includegraphics[width=\columnwidth]{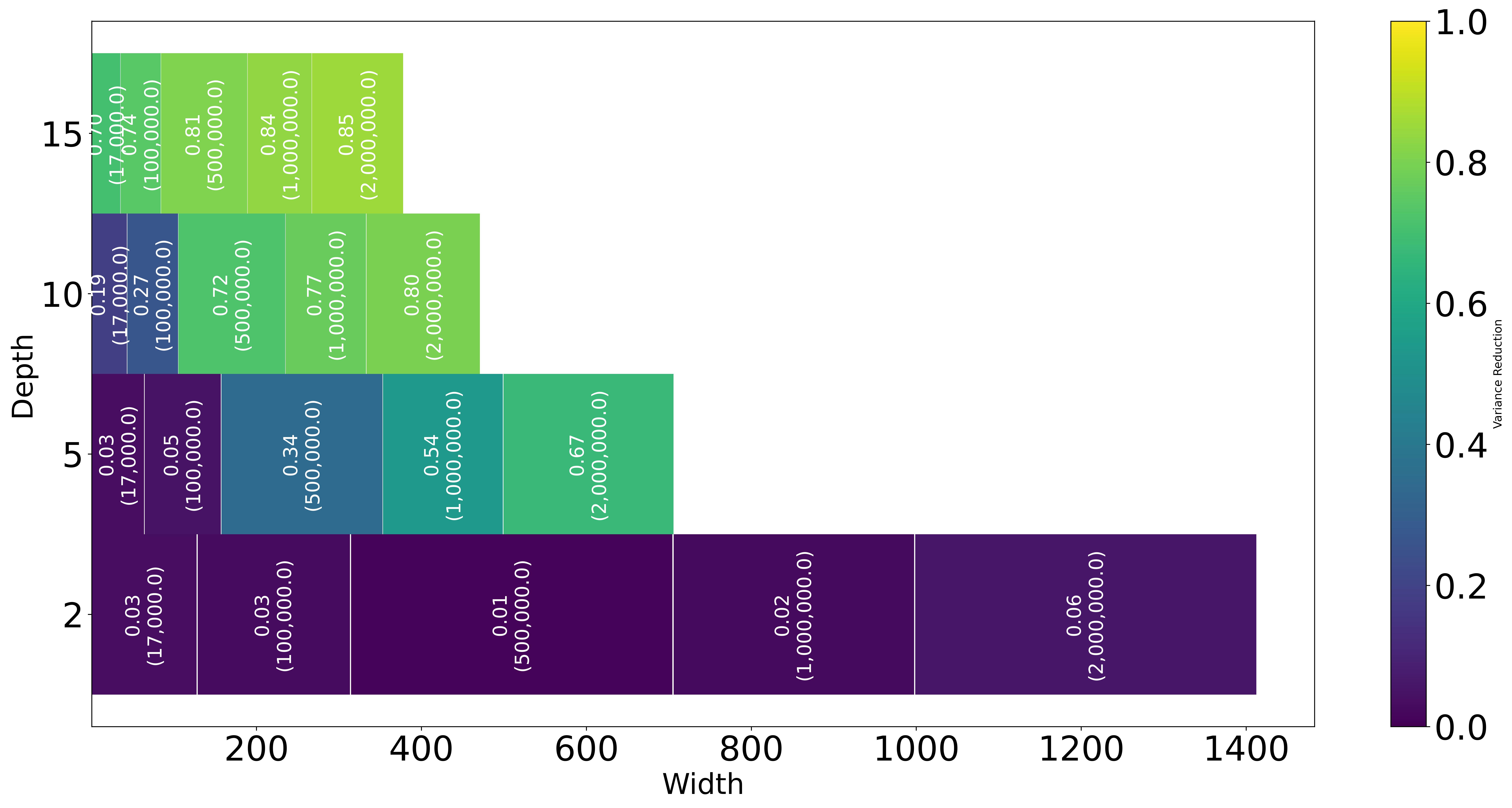} 
        \label{fig:wid_depth_spectral}
    \end{minipage}
    
    \label{fig:sidebyside}
\end{figure}

\cite{telgarsky2015representation, telgarsky2016benefits} proved that the irregularity of an MLP increases exponentially with its depth and linearly with its width. Thus, for a fixed number of parameters, a deep and narrow MLP should be harder to approximate with Taylor expansion than a shallow and wide one, and so the variance reduction should be worse for the deep network than the wide one. To test that, we report the variance reduction of various MLP with $N$ parameters, $\operatorname{MLP}_{W,D}(N)$, allocated through different width and depth combination $(W, D)$. \Cref{fig:width_depth_varred} reports that $\operatorname{MLP}_{W, D_1}(N+n)$ suffers from worse variance reduction than $\operatorname{MLP}_{W_1, D}(N+n)$, where $W_1 > W$ and $D_1 > D$. This supports \cite{telgarsky2015representation, telgarsky2016benefits}'s result, that $\operatorname{MLP}_{W, D_1}(N+n)$, is more irregular than $\operatorname{MLP}_{W_1, D}(N+n)$, hinting that increasing the irregularity of the network makes it harder to approximate with Taylor expansion; consequently reducing the variance through Taylor based control variate becomes increasingly hard when dealing with large networks with complicated transformation.

To confirm furthermore this hypothesis, we smooth the loss landscape, which should be easier to approximate, and measure the variance reduction. We apply to every linear layer of the MLP spectral normalisation \citep{miyato2018spectral}, which constrains its Lipschitz constant to one. When applied to every layer of an MLP with ReLU activation, the Lipschitz constant of the MLP is also constrained to one. 
In \cref{app:bound}, we prove that the remainder of the Taylor approximation can be bounded by the Lipschitz constant of the k-th derivative (assuming it exists), which motivates this constraint. \Cref{fig:wid_depth_spectral} reports the variance reduction of the same MLP as in \cref{fig:width_depth_varred}, and we observe improved variance reduction, which supports the hypothesis that a smoothed landscape is easier to control.

\subsection{Optimiser}
To study the effect of the optimiser, we train an MLP on the same simple setting as earlier using both Adam and SGD. As we can see in (\cref{fig:varred_mlp_sgdadam} we are able to reduce more the variance when training with SGD; because more variance is available to be reduced. The decrease in variance reduction indicates that Adam suffers less from variance in the objective.

\subsection{Discussion}
\label{sec:discussion}
We have introduced a framework to derive arbitrarily precise control variate for the training objective of a score-based model and its gradient through Taylor expansions. We have shown experimentally that in a simple controlled setting, the benefit of using the control variate to reduce the variance. Surprisingly, this benefit does not translate to the more complicated datasets and models we have tried. This un-intuitive result raises the question if the variance present in diffusion models is actually harmful to the learning objective, or actually a benefit.

\section{Conclusion}

In this study, we introduced a framework to derive arbitrarily precise control variate for the training objective of a score-based model and its gradient through Taylor expansions. In addition, we proved an equivalence between controlling the training objective and its gradients, thereby laying the foundation for future work on the relationship between reducing the variance of a training objective and its gradients. We show, theoretically and empirically that despite this equivalence, it is necessary to control the gradients variance, because of the regression coefficient that scales the control variate and allows it to take effect only when the estimator and the control variate are correlated. 
In this initial investigation we have shown that the quality of the control variate decreases with the complexity of the network \cref{sec:irregularity}, and presented evidence that higher-order expansion yields better variance reduction \cref{sec:varredmnist}. An avenue of research would be to study the relationship between $k$ and the variance reduction ratio $\rho$, for any order.\\
We also proved an equivalence between controlling the objective function and its gradients, with the equality: $\partial_{\boldsymbol{\theta}}C^{k}_{\boldsymbol{\theta}}(\mathbf{z}, \mathbf{x}, \sigma) = \mathcal{C}^{k}_{\boldsymbol{\theta}}(\mathbf{z}, \mathbf{x}, \sigma)$. However, their regression coefficient $\beta$ and $\beta_{\mathbf{g}}$ differ, which is why we can not control the gradient's variance through the objective's function. We hypothesise that \citet{wang2020wasserstein} were able to achieve it because their case happened to have $\beta = 1$, which happens when the network and the dataset are simple enough, for which most of the signal is included in the zero-th order term of the Taylor expansion. Estimating $\beta_{\mathbf{g}}$ poses a challenge, as it requires computing the variance of the gradients as well as the covariance between the gradients and their control variate. This procedure can not be efficiently addressed by automatic differentiation packages, and even though partial solutions exist \citep{dangel2020backpack}, the most effective approach remains to compute the gradients as a batch and compute any statistics from it. The drawback of this approach is the excessive memory consumption. \\

\section{Acknowldgement}
We thank Nicholas Krämer for constructive discussion regarding the project and its implementation. We acknowledge EuroHPC Joint Undertaking for awarding us access to Karolina at IT4Innovations, Czech Republic. PJ is supported by the Danish Data Science Academy, which is funded by the Novo Nordisk Foundation (NNF21SA0069429) and VILLUM FONDEN (40516).

\bibliography{main}
\bibliographystyle{icml2024}

\newpage
\appendix
\onecolumn

\section{Appendix / supplemental material}

\section{Proof}
\subsection{Control variate for small $\sigma$}
\subsubsection{Control variate on the training objective}
\label{app:cvl_small}
We provide hereafter the derivation of the control variate on the training objective.
\begin{align}
 &L_{\boldsymbol{\theta}}(\mathbf{z}, \mathbf{x}, \sigma) = \frac{1}{2}\left\| \frac{\mathbf{z}}{\sigma} + s_{\boldsymbol{\theta}}(\mathbf{x} + \sigma \mathbf{z})\right\|^2 \\
 &= \frac{1}{2} \left\|\frac{\mathbf{z}}{\sigma}\right\|^2 + \frac{1}{2} \left\| s_{\boldsymbol{\theta}}(\mathbf{x} + \sigma \mathbf{z}) \right\|^2 + \left\langle \frac{\mathbf{z}}{\sigma} | s_{\boldsymbol{\theta}}(\mathbf{x} + \sigma \mathbf{z}) \right\rangle \\
 &\simeq L^k_{\boldsymbol{\theta}}(\mathbf{z}, \mathbf{x}, \sigma) = \frac{1}{2} \left\|\frac{\mathbf{z}}{\sigma}\right\|^2 + \frac{1}{2}  \left\| T_{s_{\boldsymbol{\theta}}, \mathbf{x}}^k(\mathbf{x} + \sigma \mathbf{z}) \right\|^2 + \left\langle \frac{\mathbf{z}}{\sigma} | T_{s_{\boldsymbol{\theta}}, \mathbf{x}}^k(\mathbf{x} + \sigma \mathbf{z})  \right\rangle \\
 &=\frac{1}{2} \left\|\frac{\mathbf{z}}{\sigma}\right\|^2 + \frac{1}{2}\left\langle \sum_{|\alpha| \leq k}  \frac{\sigma^{|\boldsymbol{\alpha}|}}{ \alpha !} \mathbf{z}^{\boldsymbol{\alpha}} \partial^\alpha  s_{\boldsymbol{\theta}}(\mathbf{x}) \mid \sum_{|\rho| \leq k}  \frac{\sigma^{|\boldsymbol{\rho}|}}{ \rho !} \mathbf{z}^{\boldsymbol{\rho} } \partial^\rho  s_{\boldsymbol{\theta}}(\mathbf{x})\right\rangle + \left\langle \frac{\mathbf{z}}{\sigma} \mid \sum_{|\alpha| \leq k}  \frac{\sigma^{|\boldsymbol{\alpha}|}}{ \alpha !} \mathbf{z}^{\boldsymbol{\alpha} } \partial^\alpha  s_{\boldsymbol{\theta}}(\mathbf{x})\right\rangle\\
 &= \frac{1}{2} \left\|\frac{\mathbf{z}}{\sigma}\right\|^2 + \frac{1}{2}  \sum_{\substack{ |\alpha| \leq k \\ |\rho| \leq k}}  \frac{\sigma^{|\boldsymbol{\alpha}| + |\boldsymbol{\rho}|}}{\alpha !\rho !} \mathbf{z}^{\boldsymbol{\alpha} + \boldsymbol{\rho}} \partial^\alpha s_{\boldsymbol{\theta}}(\mathbf{x})^T\partial^\rho s_{\boldsymbol{\theta}}(\mathbf{x}) + \sum_{|\alpha| \leq k}  \frac{\sigma^{|\boldsymbol{\alpha}|-1}}{ \alpha !} \mathbf{z}^{\boldsymbol{\alpha} }\mathbf{z}^T \partial^\alpha  s_{\boldsymbol{\theta}}(\mathbf{x})
\end{align}

Taking the expectation of $L^k_{\boldsymbol{\theta}}(\mathbf{z}, \mathbf{x}, \sigma)$ quantity yields:

\begin{equation}
    \mathbb{E}_{\mathbf{z}}[L^k_{\boldsymbol{\theta}}(\mathbf{z}, \mathbf{x}, \sigma)] = \frac{1}{2} \left\|\frac{\mathbf{z}}{\sigma}\right\|^2 + \frac{1}{2}  \sum_{\substack{ |\alpha| \leq k \\ |\rho| \leq k}}  \frac{\sigma^{|\boldsymbol{\alpha}| + |\boldsymbol{\rho}|}}{\alpha !\rho !} \delta_{\boldsymbol{\alpha} + \boldsymbol{\rho}} \partial^\alpha s_{\boldsymbol{\theta}}(\mathbf{x})^T\partial^\rho s_{\boldsymbol{\theta}}(\mathbf{x}) + \sum_{|\alpha| \leq k}  \frac{\sigma^{|\boldsymbol{\alpha}|-1}}{ \alpha !} \mathbb{E}[\mathbf{z}^{\boldsymbol{\alpha} }\mathbf{z}^T] \partial^\alpha  s_{\boldsymbol{\theta}}(\mathbf{x})
\end{equation}

To conclude, we derive the control variate on the training objective as such:

\begin{equation}
\begin{aligned}
    &C^{k}_{\boldsymbol{\theta}}(\mathbf{z}, \mathbf{x}, \sigma) = L^k_{\boldsymbol{\theta}}(\mathbf{z}, \mathbf{x}, \sigma) - \mathbb{E}_{\mathbf{z}}[L^k_{\boldsymbol{\theta}}(\mathbf{z}, \mathbf{x}, \sigma)] \\
     &=\frac{\|\mathbf{z}\|^2 -D}{2 \sigma^2} +  \frac{1}{2} \sum_{\substack{|\alpha| \leq k \\|\rho| \leq k}}  \frac{\sigma^{|\boldsymbol{\alpha}| + |\boldsymbol{\rho}|}}{\alpha !\rho !} \left(\mathbf{z}^{\boldsymbol{\alpha} + \boldsymbol{\rho}} - \delta_{\boldsymbol{\alpha} + \boldsymbol{\rho}}\right)  \partial^\alpha s_{\boldsymbol{\theta}}(\mathbf{x})^T \partial^\rho s_{\boldsymbol{\theta}}(\mathbf{x}) \\
     &+ \sum_{|\alpha| \leq k}  \frac{\sigma^{|\boldsymbol{\alpha}|-1}}{ \alpha !} \left(\mathbf{z}^{\boldsymbol{\alpha}}\mathbf{z}^T -\mathbb{E}[\mathbf{z}^{\alpha}\mathbf{z}] \right) \partial^\alpha  s_{\boldsymbol{\theta}}(\mathbf{x})
\end{aligned}
\end{equation}

\subsection{Control variate for $k=1$ and $k=2$}
\label{sec:cv12}
In practice we use the control variate $C_{\textbf{g}, \boldsymbol{\theta}}^{1}(\mathbf{z}, \mathbf{x}, \sigma)$, that we get by deriving first $C_{\boldsymbol{\theta}}^{1}(\mathbf{z}, \mathbf{x}, \sigma)$ and differentiate it with automatic differentiation to get a control variate on the gradients. We also derive the control variate on the training objective for $k=2$:

\begin{align}
    C_{\boldsymbol{\theta}}^{1}(\mathbf{z}, \mathbf{x}, \sigma) &= \frac{\|\mathbf{z}\|^2 - D}{2\sigma^2} + \frac{\mathbf{z}^T}{\sigma}s(\mathbf{x}) + \mathbf{z}^T\left(\mathbf{z}^T\partial s(\mathbf{x})\right) - \operatorname{Tr}(\partial s(\mathbf{x})) \\
    & + \sigma s(\mathbf{x})^T\left(\mathbf{z}^T\partial s(\mathbf{x}) \right) + \frac{\sigma^2}{2} \left(\| \mathbf{z}^T\partial s(\mathbf{x}) \|^2  - \|\partial s(\mathbf{x})\|^2_F\right)
\end{align}
\begin{align}
    C_{\boldsymbol{\theta}}^{2}(\mathbf{z}, \mathbf{x}, \sigma) &= \frac{\|\mathbf{z}\|^2 - D}{2\sigma^2} + \frac{\mathbf{z}^T}{\sigma}s(\mathbf{x}) + \mathbf{z}^T\left(\mathbf{z}^T\partial s(\mathbf{x})\right) - \operatorname{Tr}(\partial s(\mathbf{x})) + \frac{\sigma}{2}\mathbf{z}^T\left(\mathbf{z}^T\partial s(\mathbf{x})^2\mathbf{z}\right)\\
    & + \sigma s(\mathbf{x})^T\left(\mathbf{z}^T\partial s(\mathbf{x}) \right) + \frac{\sigma^2}{2} \left(\| \mathbf{z}^T\partial s(\mathbf{x}) \|^2  - \|\partial s(\mathbf{x})\|^2_F\right) + \frac{\sigma^3}{2}\left(\mathbf{z}^T\partial(\mathbf{x})|(\mathbf{z}^T\partial s(\mathbf{x})^2\mathbf{z}\right)\\
    &+ \frac{\sigma^4}{8}\|\mathbf{z}^T\partial s(\mathbf{x})\mathbf{z}\|^2 - 2\operatorname{Tr}((\partial^2s(\mathbf{x}))^2) - \operatorname{Tr}(\partial^2s(\mathbf{x}))^2
\end{align}
\subsubsection{Control variate on the training objective's gradients}
\label{app:cvg_small}
\begin{equation}
\begin{aligned}
     \partial_{\boldsymbol{\theta}}L(\mathbf{z}, \mathbf{x}, \sigma) &= \left( \frac{\mathbf{z}}{\sigma} + s_{\boldsymbol{\theta}}(\mathbf{x} + \sigma \mathbf{z})\right)^T \partial_{\boldsymbol{\theta}}s_{\boldsymbol{\theta}}(\mathbf{x} + \sigma \mathbf{z}) \\
    &\simeq G^k(\mathbf{z}, \mathbf{x}, \sigma) = \left( \frac{\mathbf{z}}{\sigma} + T_{s_{\boldsymbol{\theta}}, \mathbf{x}}^{k}(\mathbf{x} + \sigma \mathbf{z})\right)^T  T_{ \partial_{\boldsymbol{\theta}}s_{\boldsymbol{\theta}}, \mathbf{x}}^{k}(\mathbf{x} + \sigma \mathbf{z}) \\
    &=  \left( \frac{\mathbf{z}}{\sigma} + \sum_{|\alpha| \leq k}  \frac{\sigma^{|\boldsymbol{\alpha}|}}{ \alpha !} \mathbf{z}^{\boldsymbol{\alpha}} \partial^\alpha  s_{\boldsymbol{\theta}}(\mathbf{x})\right)^T  \left( \sum_{|\rho| \leq k}  \frac{\sigma^{|\boldsymbol{\rho}|}}{ \rho !} \mathbf{z}^{\boldsymbol{\rho}} \partial^\rho  \partial_{\boldsymbol{\theta}}s_{\boldsymbol{\theta}}(\mathbf{x})\right)\\
    &= \sum_{|\rho| \leq k}  \frac{\sigma^{|\boldsymbol{\rho}|-1}}{ \rho !} \mathbf{z}^{\boldsymbol{\rho}} \mathbf{z}^T\partial^\rho  \partial_{\boldsymbol{\theta}}s_{\boldsymbol{\theta}}(\mathbf{x}) +  \sum_{\substack{|\alpha| \leq k \\ |\rho| \leq k}} \frac{\sigma^{|\boldsymbol{\alpha}| + |\boldsymbol{\rho}|  }}{\alpha! \rho!} \mathbf{z}^{\boldsymbol{\alpha} + \boldsymbol{\rho}}\partial^\alpha  s_{\boldsymbol{\theta}}(\mathbf{x})^T \partial^\rho  \partial_{\boldsymbol{\theta}}s_{\boldsymbol{\theta}}(\mathbf{x})
\end{aligned}
\end{equation}

We can now derive the expectation of $G^k(\mathbf{z}, \mathbf{x}, \sigma)$ with respect to $\mathbf{z}$ and the control variate on the gradients$\ C_{\textbf{g}, \boldsymbol{\theta}}^{k}(\mathbf{z}, \mathbf{x}, \sigma) $

\begin{equation}
    \mathbb{E}_{\mathbf{z}}[G^k(\mathbf{z}, \mathbf{x}, \sigma)] =  \sum_{|\rho| \leq k}  \frac{\sigma^{|\boldsymbol{\rho}|-1}}{ \rho !} \mathbb{E}[\mathbf{z}^{\boldsymbol{\rho}} \mathbf{z}]^T\partial^\rho  \partial_{\boldsymbol{\theta}}s_{\boldsymbol{\theta}}(\mathbf{x}) +  \sum_{\substack{|\alpha| \leq k \\ |\rho| \leq k}} \frac{\sigma^{|\boldsymbol{\alpha}| + |\boldsymbol{\rho}|  }}{\alpha! \rho!} \delta_{\boldsymbol{\alpha} + \boldsymbol{\rho}}\partial^\alpha  s_{\boldsymbol{\theta}}(\mathbf{x})^T \partial^\rho  \partial_{\boldsymbol{\theta}}s_{\boldsymbol{\theta}}(\mathbf{x})
\end{equation}
\begin{equation}
\begin{aligned}
     &C_{\textbf{g}, \boldsymbol{\theta}}^{k}(\mathbf{z}, \mathbf{x}, \sigma)  = G^k(\mathbf{z}, \mathbf{x}, \sigma) -  \mathbb{E}_{\mathbf{z}}[G^k(\mathbf{z}, \mathbf{x}, \sigma)] \\
    &= \sum_{|\rho| \leq k}  \frac{\sigma^{|\boldsymbol{\rho}|-1}}{ \rho !} \left(\mathbf{z}^{\boldsymbol{\rho}} \mathbf{z} - \mathbb{E}[\mathbf{z}^{\rho}\mathbf{z}]\right)^T\partial^\rho  \partial_{\boldsymbol{\theta}}s_{\boldsymbol{\theta}}(\mathbf{x}) + \sum_{\substack{|\rho| \leq k \\ |\alpha| \leq k}} \frac{\sigma^{|\boldsymbol{\alpha}| + |\boldsymbol{\rho}|}}{\alpha! \rho!} \left(\mathbf{z}^{\boldsymbol{\alpha} + \boldsymbol{\rho}} - \delta_{\boldsymbol{\alpha} + \boldsymbol{\rho}}\right)\partial^\alpha  s_{\boldsymbol{\theta}}(\mathbf{x})^T \partial^\rho  \partial_{\boldsymbol{\theta}}s_{\boldsymbol{\theta}}(\mathbf{x})
\end{aligned}
\end{equation}

\subsection{Proof of \Cref{thm:eqvalence}}
\label{app:proof_eqvalence}

We start by recalling the control variate on the training objective at order $k$:

\begin{equation}
\begin{aligned}
    C^{k}_{\boldsymbol{\theta}}(\mathbf{z}, \mathbf{x}, \sigma) &= \frac{\|\mathbf{z}\|^2 -D}{2 \sigma^2} +  \frac{1}{2} \sum_{\substack{|\alpha| \leq k \\|\rho| \leq k}}  \frac{\sigma^{|\boldsymbol{\alpha}| + |\boldsymbol{\rho}|}}{\alpha !\rho !} \left(\mathbf{z}^{\boldsymbol{\alpha} + \boldsymbol{\rho}} - \delta_{\boldsymbol{\alpha} + \boldsymbol{\rho}}\right)  \partial^\alpha s_{\boldsymbol{\theta}}(\mathbf{x})^T \partial^\rho s_{\boldsymbol{\theta}}(\mathbf{x}) \\
    &+ \sum_{|\alpha| \leq k}  \frac{\sigma^{|\boldsymbol{\alpha}|-1}}{ \alpha !} \left(\mathbf{z}^{\boldsymbol{\alpha}}\mathbf{z}^T -\mathbb{E}[\mathbf{z}^{\alpha}\mathbf{z}] \right) \partial^\alpha  s_{\boldsymbol{\theta}}(\mathbf{x})
\end{aligned}
\end{equation}

We take it's derivative with respect to the parameters $\mathbf{\theta}$:

\begin{equation}
\begin{aligned}
& \partial_{\boldsymbol{\theta}}C^{k}_{\boldsymbol{\theta}}(\mathbf{z}, \mathbf{x}, \sigma) = \mathbf{0} +  \frac{1}{2} \sum_{\substack{|\alpha| \leq k \\|\rho| \leq k}}  \frac{\sigma^{|\boldsymbol{\alpha}| + |\boldsymbol{\rho}|}}{\alpha !\rho !} \left(\mathbf{z}^{\boldsymbol{\alpha} + \boldsymbol{\rho}} - \delta_{\boldsymbol{\alpha} + \boldsymbol{\rho}}\right) \partial_{\boldsymbol{\theta}} \left( \partial^\alpha s_{\boldsymbol{\theta}}(\mathbf{x})^T \partial^\rho s_{\boldsymbol{\theta}}(\mathbf{x})\right) \\
&+ \sum_{|\alpha| \leq k}  \frac{\sigma^{|\boldsymbol{\alpha}|-1}}{ \alpha !} \left(\mathbf{z}^{\boldsymbol{\alpha} + 1} -\mathbb{E}[\mathbf{z}^{\alpha}\mathbf{z}] \right) \partial_{\boldsymbol{\theta}}\partial^\alpha  s_{\boldsymbol{\theta}}(\mathbf{x}) \\
&\text{We apply the product rule:} \\
& = \frac{1}{2} \sum_{\substack{|\alpha| \leq k \\|\rho| \leq k}}  \frac{\sigma^{|\boldsymbol{\alpha}| + |\boldsymbol{\rho}|}}{\alpha !\rho !} \left(\mathbf{z}^{\boldsymbol{\alpha} + \boldsymbol{\rho}} - \delta_{\boldsymbol{\alpha} + \boldsymbol{\rho}}\right) \left(\partial^\alpha  \partial_{\boldsymbol{\theta}}s_{\boldsymbol{\theta}}(\mathbf{x})^T \partial^\rho s_{\boldsymbol{\theta}}(\mathbf{x})+ \partial^\alpha s_{\boldsymbol{\theta}}(\mathbf{x})^T \partial^\rho  \partial_{\boldsymbol{\theta}}s_{\boldsymbol{\theta}}(\mathbf{x})\right) \\
&+ \sum_{|\alpha| \leq k}  \frac{\sigma^{|\boldsymbol{\alpha}|-1}}{ \alpha !} \left(\mathbf{z}^{\boldsymbol{\alpha}}\mathbf{z} - \mathbb{E}[\mathbf{z}^{\alpha}\mathbf{z}] \right)^T \partial^\alpha \partial_{\boldsymbol{\theta}} s_{\boldsymbol{\theta}}(\mathbf{x}) \\
& = \sum_{\substack{|\alpha| \leq k \\|\rho| \leq k}}  \frac{\sigma^{|\boldsymbol{\alpha}|  + |\boldsymbol{\rho}|}}{\alpha !\rho !} \left(\mathbf{z}^{\boldsymbol{\alpha} + \boldsymbol{\rho}} - \delta_{\boldsymbol{\alpha} + \boldsymbol{\rho}}\right) \left(  \partial^\alpha  \partial_{\boldsymbol{\theta}}s_{\boldsymbol{\theta}}(\mathbf{x})^T \partial^\rho s_{\boldsymbol{\theta}}(\mathbf{x})\right) \\
&+ \sum_{|\alpha| \leq k}  \frac{\sigma^{|\boldsymbol{\alpha}|-1}}{ \alpha !} \left(\mathbf{z}^{\boldsymbol{\alpha}}\mathbf{z} - \mathbb{E}[\mathbf{z}^{\alpha}\mathbf{z}]\right)^T \partial^\alpha \partial_{\boldsymbol{\theta}} s_{\boldsymbol{\theta}}(\mathbf{x}) \\
& = C^{k}_{\mathbf{g}, \boldsymbol{\theta}}(\mathbf{z}, \mathbf{x}, \sigma)
\end{aligned}
\end{equation}

Which concludes the proof.

\subsection{Control variate for large $\sigma$}
\label{app:cvg_large}
\subsubsection{Control variate on the training objective}

\begin{align}
 &L_{\boldsymbol{\theta}}(\mathbf{z}, \mathbf{x}, \sigma) = \frac{1}{2}\left\| \frac{\mathbf{z}}{\sigma} + s_{\boldsymbol{\theta}}(\mathbf{x} + \sigma \mathbf{z})\right\|^2 \\
 &= \frac{1}{2} \left\|\frac{\mathbf{z}}{\sigma}\right\|^2 + \frac{1}{2} \left\| s_{\boldsymbol{\theta}}(\mathbf{x} + \sigma \mathbf{z}) \right\|^2 + \left\langle \frac{\mathbf{z}}{\sigma} | s_{\boldsymbol{\theta}}(\mathbf{x} + \sigma \mathbf{z}) \right\rangle \\
 &\simeq \frac{1}{2}  \left\| T_{s_{\boldsymbol{\theta}}, \sigma\mathbf{z}}^k(\mathbf{x} + \sigma \mathbf{z}) \right\|^2 + \left\langle \frac{\mathbf{z}}{\sigma} | T_{s_{\boldsymbol{\theta}}, \sigma\mathbf{z}}^k(\mathbf{x} + \sigma \mathbf{z})  \right\rangle \\
 &= \frac{1}{2}  \sum_{\substack{ |\alpha| \leq k \\ |\rho| \leq k}}  \frac{\mathbf{x}^{\boldsymbol{\alpha}+ \boldsymbol{\rho}}}{\alpha !\rho !} \partial^\alpha s_{\boldsymbol{\theta}}(\sigma \mathbf{z})^T\partial^\rho s_{\boldsymbol{\theta}}(\sigma \mathbf{z}) \nonumber + \sum_{|\alpha| \leq k}  \frac{\mathbf{x}^{|\boldsymbol{\alpha}|}}{ \sigma \alpha !}\mathbf{z}^T \partial^\alpha  s_{\boldsymbol{\theta}}(\sigma \mathbf{z})
\end{align}

We take the expectation of the Taylor approximation with respect to the data $\mathbf{x}$ and subtract it to itself which yields the control variate on the training objective:

\begin{align}
\mathcal{C}_{\mathbf{\theta}}^k(\mathbf{z}, \mathbf{x}, \sigma) =  \frac{1}{2}  \sum_{\substack{ |\alpha| \leq k \\ |\rho| \leq k}}  \frac{\mathbf{x}^{\boldsymbol{\alpha} + \boldsymbol{\rho}} - \mathbf{\mu}_{\boldsymbol{\alpha} + \boldsymbol{\rho}}}{\alpha !\rho !} \partial^\alpha s_{\boldsymbol{\theta}}(\sigma \mathbf{z})^T\partial^\rho s_{\boldsymbol{\theta}}(\sigma \mathbf{z}) \nonumber + \sum_{|\alpha| \leq k}  \frac{\mathbf{x}^{|\boldsymbol{\alpha}|} - \mathbf{\mu}_{\mathbf{\alpha}}}{ \sigma \alpha !}\mathbf{z}^T \partial^\alpha  s_{\boldsymbol{\theta}}(\sigma \mathbf{z})
\end{align}
\subsubsection{Control variate on the training objective's gradients}

\begin{align}
     &\nabla_{\boldsymbol{\theta}}L(\mathbf{z}, \mathbf{x}, \sigma) =\left( \frac{\mathbf{z}}{\sigma} + s_{\boldsymbol{\theta}}(\mathbf{x} + \sigma \mathbf{z})\right)^T \partial_{\boldsymbol{\theta}}s_{\boldsymbol{\theta}}(\mathbf{x} + \sigma \mathbf{z}) \\
    &\simeq \left( \frac{\mathbf{z}}{\sigma} + T_{s_{\boldsymbol{\theta}}, \sigma\mathbf{z}}^{k_1}(\mathbf{x} + \sigma \mathbf{z})\right)^T  T_{ \partial_{\boldsymbol{\theta}}s_{\boldsymbol{\theta}}}^{k}(\mathbf{x} + \sigma \mathbf{z}) \\
    &=  \left( \frac{\mathbf{z}}{\sigma} + \sum_{|\alpha| \leq k_1}  \frac{\mathbf{x}^{\mathbf{\alpha}}}{ \alpha !} \partial^\alpha  s_{\boldsymbol{\theta}}(\sigma \mathbf{z})\right)^T  \left( \sum_{|\alpha| \leq k}  \frac{\mathbf{x}^{\boldsymbol{\rho}}}{ \rho !}  \partial^\rho  \partial_{\boldsymbol{\theta}}s_{\boldsymbol{\theta}}(\sigma \mathbf{z})\right)\\
    &=\sum_{|\alpha| \leq k}  \frac{\mathbf{x}^{\boldsymbol{\alpha}}}{ \sigma\alpha !} \mathbf{z}^T \partial^\alpha \partial_{\boldsymbol{\theta}}s_{\boldsymbol{\theta}}(\sigma \mathbf{z}) \nonumber + \sum_{\substack{|\alpha| \leq k \\ |\rho| \leq k}} \frac{\mathbf{x}^{\boldsymbol{\alpha} + \boldsymbol{\rho} }}{\alpha! \rho!} \partial^\alpha  s_{\boldsymbol{\theta}}(\sigma \mathbf{z})^T \partial^\rho  \partial_{\boldsymbol{\theta}}s_{\boldsymbol{\theta}}(\sigma \mathbf{z})
\end{align}

As before, we take the expectation with respect to the data $\mathbf{x}$ and subtract it to the approximation, which yields the $k$-th order control variate on the gradients for large value of $\sigma$:

\begin{equation}
    \mathcal{C}_{\mathbf{g}, \mathbf{\theta}}^k(\mathbf{z}, \mathbf{x}, \sigma) = \sum_{|\alpha| \leq k}  \frac{\mathbf{x}^{\boldsymbol{\alpha}} - \boldsymbol{\mu_{\alpha}}}{ \sigma\alpha !} \mathbf{z}^T \partial^\alpha \partial_{\boldsymbol{\theta}}s_{\boldsymbol{\theta}}(\sigma \mathbf{z}) \nonumber + \sum_{\substack{|\alpha| \leq k \\ |\rho| \leq k}} \frac{\mathbf{x}^{\boldsymbol{\alpha} + \boldsymbol{\rho} } - \boldsymbol{\mu_{\alpha + \rho}}}{\alpha! \rho!} \partial^\alpha  s_{\boldsymbol{\theta}}(\sigma \mathbf{z})^T \partial^\rho  \partial_{\boldsymbol{\theta}}s_{\boldsymbol{\theta}}(\sigma \mathbf{z})
\end{equation}

\section{Bounding the remainder of a Lipschitz continuous Taylor expansion}
\label{app:bound}

Let $ f: \mathbb{R}^n \rightarrow \mathbb{R}$ be a $k+1$-differentiable mapping, where each derivative $D^{k}f$ is $L_k$-Lipschitz continuous. They Taylor expansion of $f$ at $\mathbf{a} \in \mathbb{R}^n$ at order $k$ is given by:
$$f(\mathbf{x}) = T_k(\mathbf{x}) + R_k(\mathbf{x}) $$
 where 
 $$ R_k(\mathbf{x}) = \int_0^1 \frac{(1-t)^k}{k!} D^{k+1}f(\mathbf{a} + t(\mathbf{x}-\mathbf{a}))(\mathbf{x}-\mathbf{a})^{k+1} dt $$

 We bound the remainder as follows:
$$
\begin{aligned}
& \left|R_k(\mathbf{x})\right| \leq \int_0^1 \frac{(1-t)^k}{k!} L_{k+1}\|\mathbf{x}-\mathbf{a}\|^{k+1} d t \\
& \left|R_k(\mathbf{x})\right| \leq \frac{L_{k+1}}{k!}\|\mathbf{x}-\mathbf{a}\|^{k+1} \int_0^1(1-t)^k d t \\
& \left|R_k(\mathbf{x})\right| \leq \frac{L_{k+1}}{k!}\|\mathbf{x}-\mathbf{a}\|^{k+1} \cdot \frac{1}{k+1} \\
& \left|R_k(\mathbf{x})\right| \leq \frac{L_{k+1}\|\mathbf{x}-\mathbf{a}\|^{k+1}}{(k+1)!}
\end{aligned}
$$

Thus, the remainder is bounded by the Lipschitz constant of the $k+1$-derivative, which motivates constraining the Lipschitz constant of the function and its successive derivative.

\section{Experiments details}
In \cref{sec:toydatasec}, unless, specified otherwise, we train an MLP of two layers and 128 neurons per hidden layer with Adam, with a learning rate of 0.001 and the defaults parameters. \\
In \cref{sec:varredmnist}, we train a U-Net of approximately 2M parameters with Adam and its defaults parameters. The backbone is a sequence of convolutions and max pooling of $[32, 64, 128, 256, 256]$ channels. We trained the U-Net on a single A100 for $k=0$ and two A100 for $k=1$.


\end{document}